\documentclass[journal,compsoc]{IEEEtran}
\usepackage{amsmath,amsfonts}
\usepackage{algorithmic}
\usepackage{algorithm}
\usepackage{array}
\usepackage[caption=false,font=normalsize,labelfont=sf,textfont=sf]{subfig}
\usepackage{textcomp}
\usepackage{stfloats}
\usepackage{url}
\usepackage{verbatim}
\usepackage{graphicx}
\usepackage{cite}
\usepackage{makecell} 
\usepackage{capt-of}
\usepackage[pagebackref,breaklinks,colorlinks]{hyperref}
\usepackage{booktabs}
\usepackage{adjustbox}
\usepackage{multirow}
\usepackage{multirow}
\usepackage{hhline}   
\usepackage[normalem]{ulem}   
\usepackage{enumitem}
\usepackage{makecell} 
\usepackage{colortbl} 
\definecolor{lightgray}{gray}{0.9}
\definecolor{lightgreen}{RGB}{220,245,220}
\usepackage{pifont}
\newcommand{\cmark}{\ding{51}} 
\newcommand{\xmark}{\ding{55}} 

\begin{document}

\title{Action-Slot: Structured Action-Centric Representation Learning for Multi-Agent Atomic Activity Understanding}

\author{Yu-Ho Chang\textsuperscript{1}$^{\ast}$,
        Chi-Hsi Kung\textsuperscript{1}$^{\ast}$,
        Yi-Hsuan Tsai\textsuperscript{2},
        Yi-Ting Chen\textsuperscript{1}$^{\dagger}$%
\IEEEcompsocitemizethanks{
\IEEEcompsocthanksitem $^{\ast}$ These authors contributed equally to this work.
\IEEEcompsocthanksitem $^{\dagger}$ is the corresponding author.
\IEEEcompsocthanksitem \textsuperscript{1}National Yang Ming Chiao Tung University.
E-mail: \{reezzz.cs11,\,chkung,\,ychen\}@nycu.edu.tw
\IEEEcompsocthanksitem \textsuperscript{2}Atmanity Inc. E-mail: wasidennis@gmail.com
}%
}

\IEEEtitleabstractindextext{
\begin{abstract}
Atomic activity understanding aims to recognize and localize structured traffic behaviors that jointly encode motion patterns and their grounding in road topology.
Unlike conventional action recognition, atomic activities are inherently multi-agent, multi-label, and topology-aware, where multiple activities may co-occur while many agents remain inactive. This setting calls for structured action-centric representation learning, requiring decomposing a scene into semantically meaningful activity components.
We introduce Action-Slot, a structured action-centric representation learning framework. 
While slot attention has been widely used for object-centric decomposition, its permutation-invariant design and object-level inductive bias are misaligned with atomic activity semantics. 
We reformulate slot learning as structured activity decomposition through three key designs: (1) category-aligned action slots that impose semantic anchoring over predefined atomic activity categories, (2) parallel spatio-temporal slot updating for holistic video-level reasoning, and (3) background and negative-slot regularization that enforces competition between foreground activities and irrelevant regions.
Together, these designs establish an activity-centric inductive bias that disentangles concurrent and asynchronous atomic activities directly from raw video.
Beyond recognition, we show that the learned representations encode transferable spatial-temporal grounding signals. 
We further propose an attention-difference–based pseudo mask selection framework that suppresses false positives by measuring attention changes before and after candidate region removal, enabling effective weakly supervised localization.
To support systematic evaluation, we introduce TACO, a balanced dataset with full atomic activity coverage and pixel-level annotations.
Extensive experiments across synthetic and real-world datasets demonstrate superior recognition performance, cross-domain generalization, structured decomposition analysis, and weakly supervised
localization. 
Overall, this work establishes a principled action-centric representation learning framework that unifies recognition and spatial reasoning for multi-agent atomic activity understanding.

\end{abstract}

}

\maketitle

\IEEEdisplaynontitleabstractindextext
\IEEEpeerreviewmaketitle

\IEEEdisplaynontitleabstractindextext\maketitle
\IEEEpeerreviewmaketitle

\label{sec:intro}
\begin{figure*}[t!]
\centering
        \includegraphics[width=1.0\textwidth,trim={0 3cm 0 3cm},clip]{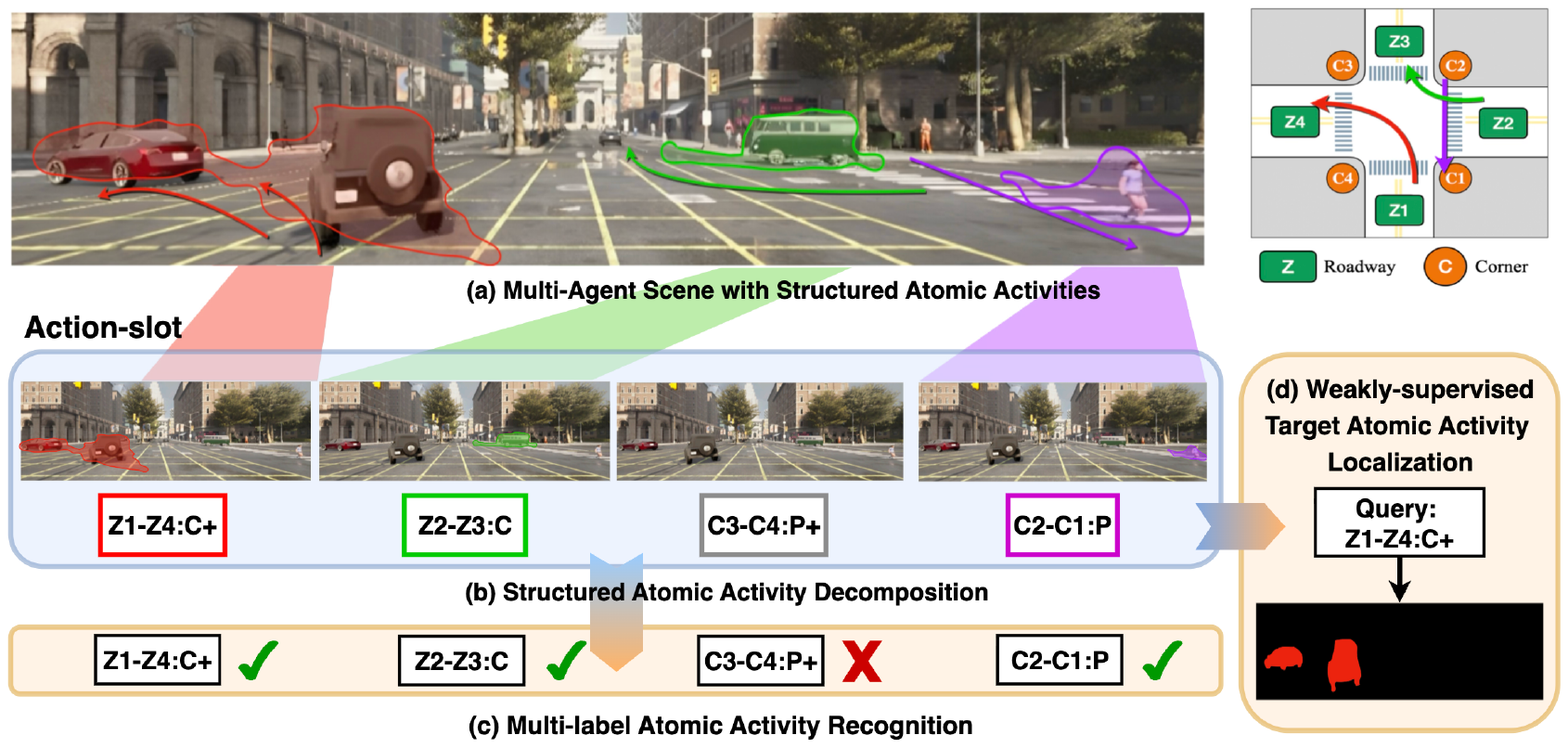}
        \captionof{figure}{
\textbf{Overview of Action-Slot for Multi-Agent Atomic Activity Understanding.}
(a) A multi-agent traffic scene containing multiple concurrent atomic activities, illustrated by colored trajectories. For example, the red trajectory corresponds to the atomic activity \textbf{Z1-Z4: C+}, representing a group of vehicles turning left along a topologically defined path. 
(b) Action-Slot decomposes the scene into semantically aligned action slots, each capturing one atomic activity without relying on object proposals.
(c) The resulting slot representations enable multi-label atomic activity recognition by predicting whether each atomic activity category is present in the scene.
(d) The learned action slots provide spatial grounding signals that enable high-quality pseudo-mask generation for weakly supervised target atomic activity localization.
%
%
        }
    \label{fig:teaser}
\end{figure*}

\section{Introduction}
\IEEEPARstart{U}{nderstanding} who is doing what, where, and when in complex dynamic scenes remains a fundamental challenge in computer vision. In multi-agent environments such as traffic intersections, multiple agents simultaneously execute distinct motion patterns under shared structural constraints such as road topology and traffic rules. Accurate recognition and localization of such structured activities are essential for downstream applications, including intention prediction~\cite{brain4car2015,stipicra2020,semanticregioncorl2022}, scenario retrieval~\cite{lin2014visual,driverwacv2020,Segal_universal_corl2020,Naphade23AIC23}, and scenario-based safety evaluation~\cite{carlachallenge2022,xu2022safebench,kung2023riskbench}.

Atomic activity, introduced in OATS~\cite{Agarwal_2023_ICCV}, provides a structured abstraction of traffic behavior by grounding motion patterns in the underlying road topology. Unlike conventional action recognition, atomic activity understanding is inherently multi-agent, multi-label, and topology-aware, where multiple activities may co-occur within a single scene, while many agents remain inactive. 
Figure~\ref{fig:teaser} illustrates the central challenge and motivation of this work. As shown in Fig.~\ref{fig:teaser} (a), a traffic scene may contain multiple concurrent atomic activities, as well as inactive objects such as the standing pedestrian at the far-right corner. 
Rather than encoding such complex dynamics with a compact feature, we argue that 
%
%
effective atomic activity understanding must decompose a scene into semantically meaningful activity components, selectively attend to relevant spatio-temporal regions, and suppress irrelevant agents and background clutter. 
%

Existing approaches are not well aligned with these structural requirements. Video-level models based on 3D ConvNets~\cite{carreira2017quo,tran2019video,feichtenhofer2019slowfast,feichtenhofer2020x3d} or the Transformer~\cite{fan2021multiscale,tong2022videomae,zhao2022tuber,gritsenko2024end} aggregate global features but lack explicit mechanisms for disentangling concurrent activities. Object-aware methods~\cite{Baradel_2018_ECCV,CVPR2019_ARG,pan2021actor,arnab2022beyond,Agarwal_2023_ICCV} leverage proposal-based representations~\cite{he2017mask,girshick2015fast}, yet they rely on explicit object supervision and primarily encode object identity rather than activity semantics grounded in topology. These limitations reveal a gap: while object-centric decomposition has been extensively studied, activity-centric decomposition remains underexplored.
This motivates a central question: \textit{Can we learn semantically structured, action-centric representations that decompose complex multi-agent activities without relying on object proposals?}

To address this question, we propose \textbf{Action-Slot}, a structured action-centric representation learning framework that reformulates slot attention as a mechanism for \textit{structured activity decomposition}.
Slot attention~\cite{locatello2020object} provides a mechanism for extracting object-centric representations.
%
%
However, directly applying it to atomic activities exposes three fundamental mismatches:
(1) permutation invariance conflicts with fixed semantic activity categories;
(2) frame-wise recurrent updates limit global temporal reasoning; and
(3) object-centric inductive biases do not align with higher-order activity semantics.

We introduce three principled designs. 
%
First, \emph{category-aligned action slots} associate each slot with a predefined atomic activity category, transforming permutation-invariant slots into semantically anchored activity components. 
Second, \emph{parallel spatio-temporal updating} enables holistic reasoning across the entire video rather than locally confined frame-level updates.
Third, \emph{background and negative-slot regularization} introduces explicit competition between foreground activity slots and background. This design enforces slot learning not attending to inactive regions. 
%
Collectively, these designs establish a new action-centric inductive bias for atomic activity understanding, as conceptualized in Fig.~\ref{fig:teaser} (b).
%

The learned representations encode spatial grounding signals beyond recognition. As illustrated in Fig.~\ref{fig:teaser} (d), we extend Action-Slot to weakly supervised target atomic activity localization, where only video-level labels are available. Raw attention maps are often diffuse and contain false positives. We therefore propose an attention-difference–based pseudo mask selection framework that measures attention changes before and after candidate region removal to identify discriminative regions. The resulting selected pseudo masks provide effective supervision for localization.
%

To facilitate systematic evaluation of atomic activity understanding, we introduce Traffic Activity Recognition Dataset (\textbf{TACO}), a large-scale synthetic dataset constructed using CARLA~\cite{Dosovitskiy17}. 
TACO provides balanced coverage of all defined atomic activity categories, including rare atomic activities, and includes pixel-level annotations for localization evaluation. 
It addresses two major gaps in OATS: class imbalance and the absence of dense spatial annotations. 
We further annotate atomic activities on nuScenes~\cite{nuscenes} to assess cross-domain generalization from synthetic to real-world data.

Extensive experiments validate Action-Slot from multiple perspectives. First, it achieves superior recognition performance on both OATS~\cite{Agarwal_2023_ICCV} and TACO. 
Second, we show that pretraining on TACO improves cross-domain performance on a real-world dataset, and that validates the effectiveness of the collected large-scale, cost-efficient synthetic data and the learned action-centric representations.
Third, we prove that action-centric decomposition is superior to explicit object-level guidance in capturing concurrent atomic activities.  
Finally, supervision with refined pseudo masks yields state-of-the-art weakly supervised target atomic activity localization performance~\cite{chen2023weakly,duan2024mining}.

Overall, this work positions action-centric representation learning as a general framework for atomic activity understanding, bridging recognition and spatial reasoning under a unified representation paradigm. 
This work makes the following contributions:
\begin{enumerate}
\item We reformulate slot attention as structured activity decomposition via category-aligned action slots, parallel spatio-temporal updating, and background and negative-
slot regularization for atomic activity understanding.
\item We unify recognition and weakly supervised target atomic activity localization through an attention-difference–based pseudo mask selection framework that generates reliable pseudo masks for supervision.
\item We establish TACO, a balanced and topology-aware benchmark with full atomic activity coverage and pixel-level annotations, enabling systematic evaluation of recognition and localization.
\item We demonstrate strong cross-domain generalization across synthetic and real-world datasets, and model architectures that validate the robustness and transferability of action-centric representations. 
\end{enumerate}

This journal version substantially expands the CVPR 2024 paper~\cite{kung2023action}. 
Beyond recognition, we generalize Action-Slot into a structured action-centric representation learning framework, introduce dense annotations enabling localization evaluation, and provide detailed analysis of weakly supervised target atomic activity localization that are not explored in the original conference version. 
Our source code, dataset, and visualization videos are available at \href{https://hcis-lab.github.io/Action-slot-TPAMI_project_page/}{https://hcis-lab.github.io/Action-slot-TPAMI\_project\_page/}.

\label{sec:related_work}
\section{Related Work}

\subsection{Video Action Recognition}
Recent advances in video action recognition have been driven by increasingly powerful spatio-temporal architectures, including 3D ConvNets~\cite{carreira2017quo,feichtenhofer2020x3d,tran2019video,feichtenhofer2019slowfast} and Transformer-based models~\cite{bertasius2021space,arnab2021vivit,fan2021multiscale,tong2022videomae,counterfactual_ICCV_2025}. Large-scale datasets such as Kinetics~\cite{kay2017kinetics} have enabled effective pretraining, significantly improving transfer performance on downstream benchmarks, including HMDB~\cite{kuehne2011hmdb}, THUMOS~\cite{THUMOS14}, and Charades~\cite{sigurdsson2016hollywood} datasets.
Beyond single-label recognition, the community has increasingly focused on multi-label action recognition, supported by benchmarks such as MultiTHUMOS~\cite{yeung2015every}, Charades~\cite{sigurdsson2016hollywood}, and AVA~\cite{gu2018ava}. Among related tasks, group activity recognition~\cite{choi2009they,volleyball} is particularly relevant, as it models multiple agents performing distinct actions within a shared scene.

Atomic activity recognition in traffic environments differs in several fundamental aspects. First, traffic scenes exhibit extreme activity sparsity: most objects remain inactive, forming an explicit negative class that must be modeled rather than implicitly ignored. This sparsity characteristic is largely absent in conventional group activity datasets. Second, atomic activities are tightly grounded in the underlying road topology, requiring joint reasoning over object identities, spatial configurations, and motion patterns under shared environmental constraints. Third, existing spatio-temporal backbones primarily rely on holistic feature aggregation over the entire video and lack explicit mechanisms for structured activity decomposition under such sparsity constraints. As a result, they struggle to disentangle multiple concurrent atomic activities while suppressing inactive regions.

A natural alternative is object-centric decomposition~\cite{Baradel_2018_ECCV,CVPR2019_ARG,pan2021actor,arnab2022beyond,Agarwal_2023_ICCV}. However, object-centric representations partition scenes according to appearance or identity rather than activity semantics. Therefore, object-level decomposition does not directly yield activity decomposition.
%
%
Slot attention provides a promising foundation and motivates our reformulation toward structured activity decomposition.

\subsection{Representation Learning via Slot Attention}
Slot attention~\cite{locatello2020object} is an unsupervised framework that is designed to discover object-centric abstractions. The initial work demonstrates its capability on synthetic datasets~\cite{johnson2017clevr}. Subsequent works have extended slot-based modeling to real-world dynamic scenes~\cite{sun2020scalability,Geiger2012CVPR}. These works incorporates auxiliary signals such as optical flow~\cite{kipf2022conditional}, motion segmentation~\cite{bao2022discovering}, object locations~\cite{elsayed2022savi++}, or depth~\cite{elsayed2022savi++} to stabilize object discovery and temporal tracking.

Extending slot attention to learn action-centric representations for atomic activity poses several challenges.
First, slots in the original formulation are permutation-invariant, which is desirable for unsupervised object discovery~\cite{locatello2020object,kipf2022conditional,elsayed2022savi++,bao2022discovering,zhou2022slot,NEURIPS2022_3dc83fcf,biza2023invariant}. However, atomic activity recognition relies on a fixed and semantically defined activity vocabulary, where permutation invariance hinders consistent alignment between slots and predefined categories.
Second, existing video-based slot architectures, such as SAVi~\cite{kipf2022conditional} and Slot-VPS~\cite{zhou2022slot}, are primarily designed for object tracking and segmentation. These methods update slots recurrently on a per-frame basis. In contrast, atomic activity understanding requires holistic temporal reasoning.
Third, existing extensions leverage object-level cues (e.g., flow or depth) to stabilize slot learning. These cues are misaligned with activity semantics: many traffic participants are inactive (e.g., pedestrians strolling
on the sidewalk and vehicles waiting at traffic lights), and motion alone does not inform atomic activity. As a result, slot learning may be biased toward saliency rather than structured activity decomposition.

To overcome these limitations, 
%
we reformulate slot binding from object instances to activity semantics, introduce parallel spatio-temporal updating for global temporal reasoning, and incorporate background-aware regularization to handle activity sparsity.
To the best of our knowledge, this work is among the first to systematically reformulate slot attention for structured activity decomposition and demonstrate the effectiveness in complex traffic scenarios.

\subsection{Weakly-Supervised Action Localization}
Recent advances in weakly supervised action segmentation have shown that meaningful localization can emerge from pretrained visual representations and auxiliary structural cues, even without access to pixel-level annotations. 
For example, VSCR~\cite{duan2024mining} combines heterogeneous contrastive cues, including CAMs, edges, motion, and semantic tags, to provide indirect supervision.
More broadly, weakly supervised action localization and segmentation methods commonly follow a synthesize-and-refine paradigm~\cite{ahn2018learning,chen2020weakly,wu2021embedded,sun2021ecs}, where pseudo masks generated from superpixels~\cite{achanta2012slic} or Grad-CAM~\cite{selvaraju2017grad} are progressively refined.
For example, Yan \textit{et al.}~\cite{yan2017weakly} employ ranking SVMs together with conditional random fields to identify representative supervoxels, while Chen \textit{et al.}~\cite{chen2020learning} introduce a cut-and-paste strategy that preserves pseudo masks whose semantics remain consistent under a discriminator.

Although effective for general action understanding benchmarks~\cite{xu2015can}, these approaches are less suitable for atomic activity recognition, where activity semantics are strongly coupled with spatial configuration and scene context. 
For instance, \textbf{C4-C1:P} and \textbf{C3-C2:P} both describe a pedestrian moving in the same direction, yet correspond to distinct activities due to their relative positions within the intersection. 
Consequently, region-level perturbations such as cut-and-paste operations may alter the underlying atomic activity label and introduce ambiguous supervision signals.
%
%
Moreover, these approaches rely on recognition models producing sufficiently discriminative attention maps to bootstrap segmentation learning. However, subtle spatial interactions in atomic activities often yield coarse and noisy activations. Without further verification, these methods struggle to effectively correct spurious responses, leading to unreliable pseudo masks.

To address this, we exploit learned action-centric representations~\cite{kung2023action} to induce activity-aligned attention maps. To further eliminate residual spurious responses, we propose an attention-difference-based selection framework that evaluates each region's causal effect by measuring attention variations upon candidate removal.

\begin{table*}[t!]
\centering
\caption{
Comparison of topology-aware traffic activity datasets. We introduce TACO, which provides balanced atomic activity categories with spatio-temporal localization annotations. The symbol $^\dagger$ denotes that we extend nuScenes with atomic activity recognition annotation. 
}
\begin{tabular}{lccccccc}
\toprule
Dataset & \# Videos & \# Labeled Activities & \# Classes & Activity Labels 
& \makecell{Spatio-Temporal Localization\\for Atomic Activity} \\
\midrule
Inner-City~\cite{chen2016atomic} & 59000 (frames) & 3.1k & 25 & High-level Action \& Location & \xmark \\
LOKI~\cite{lokiiccv2021} & 644 & \textbf{28k} & 14 & Intention & \xmark \\
ROAD~\cite{singh2022road} & 22 & 10.8k & 43 & High-level Action \& Location & \xmark \\
ROAD-Waymo~\cite{khan2024road} & 8.5 (hours) & 39k & 43 & High-level Action \& Location & \xmark \\
OATS~\cite{Agarwal_2023_ICCV}  & 1k & 6.6k & 59  & Atomic Activity & \xmark \\
ATARS~\cite{chen2025atars} & 39 & 235k (frames) & 38  & Atomic Activity & \xmark \\
\midrule
nuScenes$^{\dagger}$~\cite{nuscenes} & 426 & 933 & 51 & Atomic Activity & \xmark \\
\textbf{TACO (Ours)}  & \textbf{6.8k} & 16.5k & \textbf{64} & Atomic Activity & \cmark \\
\bottomrule
\end{tabular}
\label{tab:datasets}
\end{table*}

\subsection{Weakly-suervised Video Object Grounding}
Weakly-Supervised Video Object Grounding (WSVOG) aims to localize objects based on natural language descriptions without access to spatial annotations~\cite{Yu-IJCV-2017,huang-buch-2018-finding-it}.
A common approach in WSVOG formulates the task as a frame-level multiple instance learning (MIL) problem~\cite{zhou2018weakly, shi2019not, chen2023weakly, wang2022weakly}.
MIL-based grounding methods treat spatial proposals as instances within a positive bag and encourage at least one instance to match the supervisory signal. 
However, under activity sparsity, a video may contain numerous inactive regions that correlate with contextual cues, leading to unstable atomic activity localization.
In contrast, the proposed attention-difference
pseudo framework exploits Action-Slot’s activity-aligned attention maps to selected pseudo masks, yielding substantially improved localization performance over MIL-based approaches. 
%
%
%
%

\subsection{Traffic Activity Understanding Datasets}
Traffic activity understanding supports the development of intelligent driving systems, such as scene analysis~\cite{ramanishka2018toward,stipicra2020,semanticregioncorl2022}, scenario retrieval~\cite{lin2014visual,driverwacv2020,Segal_universal_corl2020,Naphade23AIC23}, and safety-critical scenario generation~\cite{xu2022safebench,rempe2022strive,cao2022advdo,rempeluo2023tracepace,chen2025controllable}.
Early traffic datasets~\cite{chen2016atomic,ramanishka2018toward,li2019dbus,malla2020titan} primarily annotate high-level actions (e.g., left turn) for behavior recognition.
However, such coarse labels fail to differentiate actions originating from distinct spatial contexts. For example, a left turn from the oncoming lane conveys different semantics than a left turn initiated from the ego lane.
To address this limitation, datasets such as Inner-City~\cite{chen2016atomic}, LOKI~\cite{lokiiccv2021}, ROAD~\cite{singh2022road} augment activity with location-aware (e.g., middle of the road) annotations.
Recently, OATS~\cite{Agarwal_2023_ICCV} unifies action and spatial context through topology-aware description language called \textit{Atomic Activity}, which decomposes interactive traffic scenarios into atomic activities defined by road-user types and motion patterns tied to road structures.
%
%
These datasets involve topology-aware semantics and require explicit modeling of activity sparsity.
%
%
%
%
Despite these advances, collecting real-world traffic data with comprehensive and balanced atomic activity distributions remains challenging, particularly for rare events. This limitation hinders systematic evaluation of atomic activity recognition and localization.

To overcome these challenges, we construct the TACO dataset utilizing the CARLA simulator~\cite{Dosovitskiy17} to gather instances of all conceivable activity categories, ensuring a well-balanced distribution, and further provide human-labeled spatio-temporal annotations for localization evaluation. 
%
Table~\ref{tab:datasets} compares TACO and the extended nuScenes with existing traffic activity datasets with topology-aware annotations.
By establishing a controlled and open-sourced benchmark for topology-aware atomic activity recognition and localization, TACO facilitates systematic evaluation of structured representation learning methods and encourages future research on scalable auto-labeling strategies~\cite{chen2024masktrack, ravi2024sam2segmentimages}.

\label{sec:method}

\begin{figure*}[t!]
\centering
    \includegraphics[width=17cm]{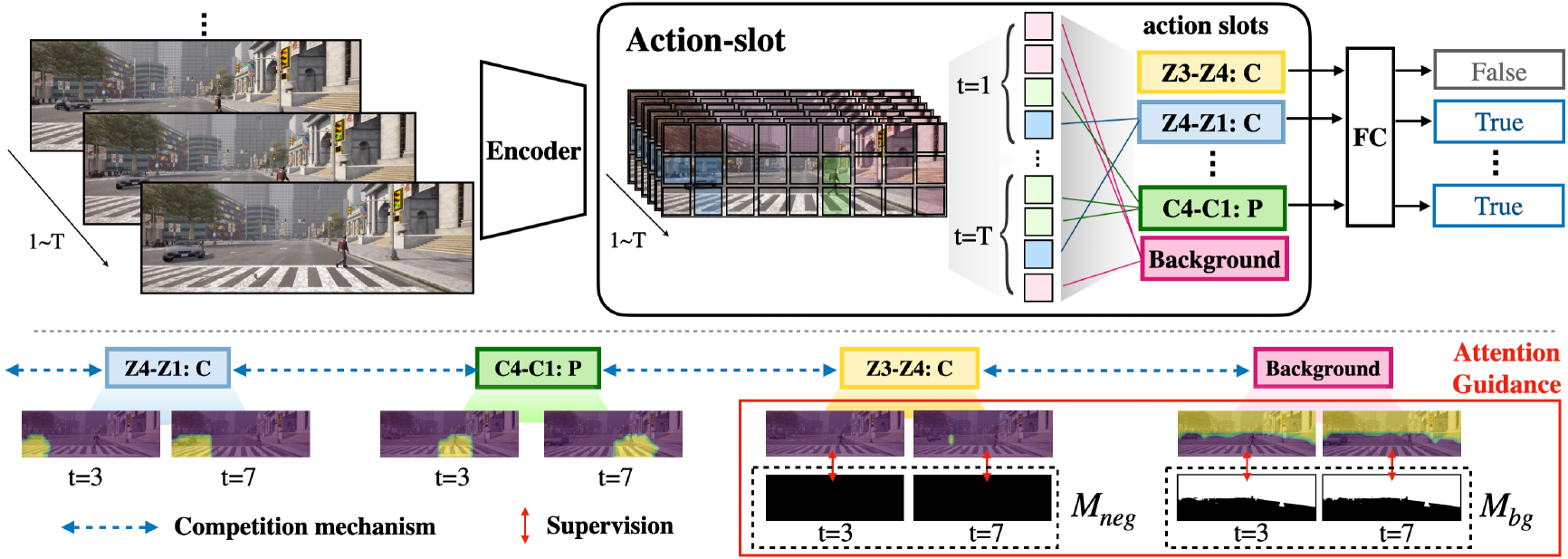}
        \caption{The top of the figure illustrates the proposed framework. Action-Slot takes video as input and uses a CNN encoder to extract feature patches. All patches are then processed with individual slots simultaneously to find the most relevant spatio-temporal patches corresponding to each action slot. The updated action slots are fed into a fully connected layer to predict the probability of the corresponding action class, excluding the background slot. The bottom of the figure depicts the attention maps of action and background slots. We propose to incorporate a background mask $M_{\mathtt{bg}}$ to supervise the background slot. The design facilitates other action slots to capture action signals. Furthermore, we design a regularization for slots allocated to negative classes (e.g., Z3-Z4: C) using an all-zero mask $M_{\mathtt{neg}}$. 
        }
        \label{fig:main_arch}
\end{figure*}

\section{Method}
In this section, we first introduce the background of Slot Attention. We then present an overview of learning action-centric representations for multi-agent atomic activity recognition using our proposed method, Action-Slot, and describe the specific modifications made to adapt Slot Attention to our task. Finally, we introduce a pseudo mask selection framework that enables weakly supervised, query-based atomic activity localization within Action-Slot.

\subsection{Preliminary: Slot Attention}
\label{sec:slot_attention}
The Slot Attention module~\cite{locatello2020object} can be viewed as a clustering algorithm that maps image patches input to a set of $K$ output slots $S \in \mathbb{R}^{K \times D_{\mathtt{slot}}}$, where $D_{\mathtt{slot}}$ is the dimension of each slot.
Specifically, the slots $S$ are first initialized by randomly sampling $K$ vectors from a Gaussian distribution, where the parameter $K$ is usually defined as the maximum number of objects in an image.
An input frame, i.e., image features $F \in$ $\mathbb{R}^{H \times W \times D_{\mathtt{in}}}$ with size $H \times W$ and dimension $D_{\mathtt{in}}$, is first flattened to patch tokens $F^{'} \in$ $\mathbb{R}^{N \times D_{\mathtt{in}}}$, where $N = H \times W$. Then the tokens are mapped to the slots using the dot product attention module.
That is, the attention weight can be calculated as $A = \frac{1}{\sqrt{D}} k(F^{'}) \cdot q(S) \in \mathbb{R}^{N \times K}$, and $q$ and $k$ are linear transformations that map the input $F$ and slots $S$ to a common dimension $D$.

We obtain the updated values for slots through $U = \bar{A}^{T} \cdot v(F^{'}) \in \mathbb{R}^{K \times D}$, where $\bar{A}$ is the normalized attention weight calculated via the softmax operation and $v$ is a linear transformation. 
Finally, the slots are updated via a GRU: $S^{'} = \mathtt{GRU}(S, U)$~\cite{cho-etal-2014-learning}. To refine the slots, the updating process repeats $M$ iterations~\cite{locatello2020object,kipf2022conditional,elsayed2022savi++}.
Note that the key difference between classical attention~\cite{vaswani2017attention} and slot attention is that the attention weights $A$ are normalized with the softmax operation slot-wise instead of token-wise. 
This difference enables slots to compete with each other so that each slot attends to different relevant regions of input. 
To extend slot attention to video tasks, prior work~\cite{kipf2022conditional,elsayed2022savi++,bao2022discovering} propagates slots recurrently via a GRU, i.e., $S^{t} = \mathtt{GRU}(S^{t-1}, U)$. 

\subsection{Overview of Proposed Method for Recognition}
\label{sec:overview}

\subsubsection{Problem Formulation}
We supervise the model to learn action-centric representations through a multi-label atomic activity recognition task. 
Specifically, given a video clip $V_i$ consisting of $T$ frames $\{I_t^i\}_{t=1}^T$, the objective is to determine whether any atomic activities $Y$ are present in the clip. The ground-truth label $Y^i$ for video $V_i$ is a binary multi-label vector, i.e., $Y^i= \{y_c\}_{c=1}^{N_{\mathtt{cl}}}$, where $y_c = 1$ if the corresponding activity appears in the video and $0$ otherwise, and $N_{\mathtt{cl}}$ denotes the total number of activity classes. 
Notably, we do not impose temporal constraints on when an activity occurs; an activity $y_c$ may appear in any frame of $\{I_t^i\}_{t=1}^T$and may occur multiple times within a single video.

\subsubsection{Overview of Action-Slot}
We propose Action-Slot, an action-centric, Slot-Attention-based framework for decomposing atomic activities in videos. Our key idea is to associate each slot with a specific atomic activity, enabling it to learn a dedicated action-centric representation; we refer to these as \emph{action slots}. 
An overview of Action-Slot is illustrated in Figure~\ref{fig:main_arch}.
Unlike existing Slot Attention approaches designed for unsupervised object discovery and tracking~\cite{kipf2022conditional,elsayed2022savi++,bao2022discovering}, we introduce three key modifications.
First, we allocate a fixed number of slots $K$ equal to the number of atomic activity classes, $N_{\mathtt{cl}}$. 
Each action slot is supervised using the corresponding ground-truth label $y_c$, indicating whether activity $c$ is present in the video.
Second, we introduce an additional background slot to capture regions unlikely to contain any activity. 
By absorbing attention over irrelevant regions, the background slot allows the action slots to focus on activity-relevant areas.
Third, we explicitly discourage inactive action slots—corresponding to activities absent from the video—from attending to any regions.
Finally, instead of recurrently propagating slots over time as in prior work~\cite{kipf2022conditional,elsayed2022savi++,bao2022discovering}, we update all action slots in parallel by jointly considering all frames.

\subsection{Action-Slot}
\label{sec:action_slot}
We now describe the details of our Action-Slot design for learning action-centric representations. Following~\cite{bao2022discovering}, we initialize slots using learnable parameters.
We extract image features
Given a video clip, we first extract frame-level features
$F \in \mathbb{R}^{T \times H \times W \times D_{in}}$ using video encoder~\cite{he2016deep,carreira2017quo,feichtenhofer2019slowfast,feichtenhofer2020x3d}
The features are then flattened into $F^{'} \in \mathbb{R}^{N \times D_{\mathtt{in}}}$, where $N = T \times H \times W$.
We further incorporate learnable 3D spatio-temporal positional embeddings $E \in \mathbb{R}^{T \times H \times W} $ into the tokens $F^{'}$~\cite{bertasius2021space,arnab2021vivit,fan2021multiscale,bao2022discovering,tong2022videomae}.

\subsubsection{Allocated Slot}
We define a set of slots $\{S_k\}_{k=1}^{K}$, where $K = N_{\mathtt{cl}}$ denotes the number of atomic activity classes.
Unlike prior Slot Attention approaches~\cite{locatello2020object,kipf2022conditional,elsayed2022savi++,zhou2022slot}, which do not assign predefined semantics to individual slots and typically rely on Hungarian matching for bipartite alignment with objects, we explicitly associate each slot with a specific activity class, forming action slots.
Each action slot is equipped with an independent binary classifier that predicts $\hat{y}_c$ for the corresponding activity class $c$. 
We supervise these predictions using a binary cross-entropy (BCE) loss for each sample: $L_{\mathtt{act}} = \sum_{c=1}^{N_{\mathtt{cl}}} \textnormal{BCE}(\hat{y}_c, y_c)$.

\subsubsection{Parallel Updating}
\begin{figure}[t!]
\centering
    \includegraphics[width=8cm]{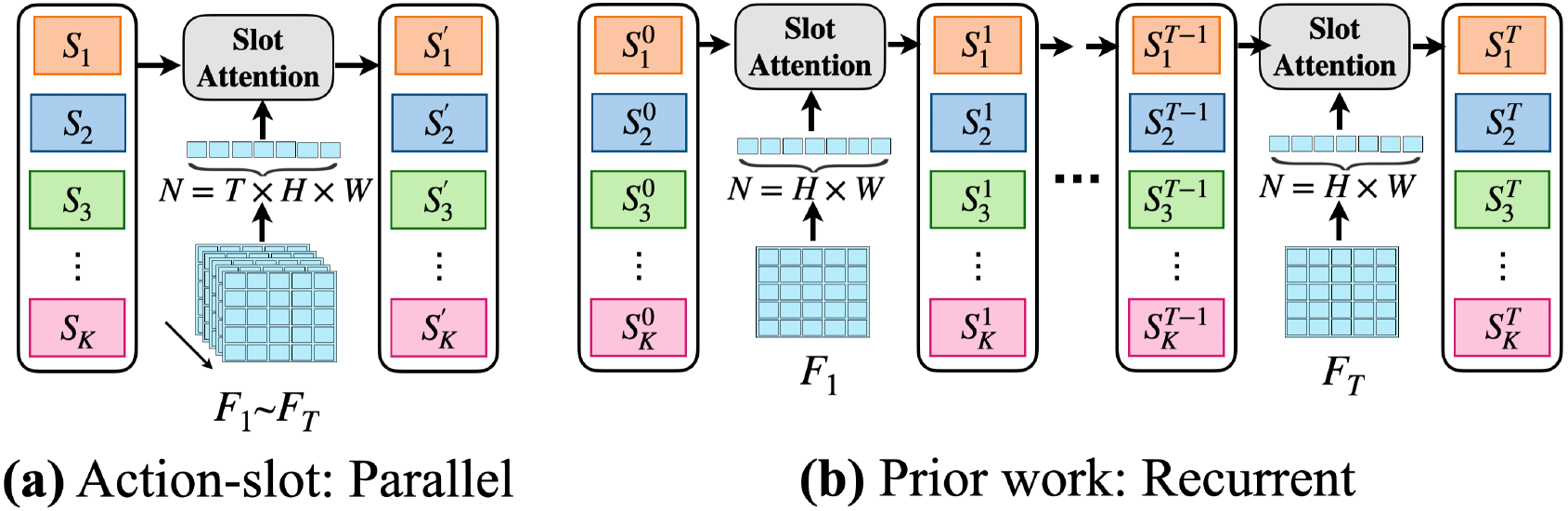}
        \vspace{-1mm}
        \caption{
       (a) Our parallel scheme considers updating slots based on the spatio-temporal features for all the frames.
       (b) The slots are updated recurrently along the temporal dimension, where each time the slot only considers a frame-wise feature.
        }
        \label{fig:parallel}
        \vspace{-4mm}
\end{figure}

We update slots across image frames in a parallel manner, as shown in Figure~\ref{fig:parallel} (a). Specifically, we compute normalized attention weight $\bar{A} \in \mathbb{R}^{N \times K}$ over the entire spatio-temporal volume, where $N = T \times H \times W$. 
The $K$ slots are then updated in a single pass using $\bar{A}$.
This differs from prior approaches\cite{kipf2022conditional,elsayed2022savi++,bao2022discovering}, which update slots recurrently across frames.
In those methods, slots are propagated over $T$ iterations, where the previous slot states and the current frame features are used to compute the updated slot states at each time step, as illustrated in Figure~\ref{fig:parallel} (b).

\subsubsection{Background Slot and Attention Guidance}
Providing direct supervision to action slots corresponding to positive classes (i.e., present atomic activities) is non-trivial, since objects are not continuously involved in activities and not all objects participate in actions.
Therefore, we propose to use a background slot that is not supervised by any activity classes. 
To address this, we introduce a background slot that is not associated with any activity class. Due to the softmax normalization in Slot Attention, the background slot is encouraged to attend to regions irrelevant to activities, thereby allowing action slots to focus on activity-relevant areas (see the bottom of Figure~\ref{fig:main_arch}).
To further strengthen this mechanism, we supervise the background slot with a background mask $M_{\mathtt{bg}}$, which explicitly guides it toward non-activity regions. 
The mask is constructed by excluding pixels belonging to vehicles, pedestrians, drivable areas, crosswalks, and sidewalks. 
The background attention is optimized using a binary cross-entropy loss:
$L_{\mathtt{bg}} = \textnormal{BCE}(\bar{A}_{\mathtt{bg}}, M_{\mathtt{bg}})$, where $\bar{A}_{\mathtt{bg}}$ denotes the normalized attention map of the background slot.

\subsubsection{Regularization for Action-Slot}
\label{subsec:reg}
As discussed earlier, directly guiding the attention of slots corresponding to positive classes is non-trivial. Instead, we introduce a regularization term that discourages action slots associated with negative classes (i.e., activities absent from the video) from attending to any regions.
To this end, we construct a negative mask $M_{\mathtt{neg}}$ whose elements are all zeros. 
We then define the following loss:
$L_{\mathtt{neg}} = \sum_{\{c | y_{c}=0\}} \textnormal{BCE}(\bar{A}_{c}, M_{\mathtt{neg}})$,
where $\bar{A}_{c}$ 
denotes the normalized attention output of slot $S_c$ for negative class $c$.
This regularization suppresses attention from inactive action slots. 
Through the inherent competition induced by the softmax operation in Slot Attention, suppressing negative slots encourages slots associated with positive classes to focus more strongly on informative regions.
Importantly, this regularization does not require any additional annotations.
The overall training objective of Action-Slot is given by
\[L_{\mathtt{all}} = L_{\mathtt{act}}+ w_{\mathtt{bg}}L_{\mathtt{bg}} + w_{\mathtt{neg}}L_{\mathtt{neg}}\], 
where $w_{\mathtt{bg}}$ and $w_{\mathtt{neg}}$ balance the background and negative regularization terms (set to 0.5 and 1.0, respectively in our experiments).

\subsection{Weakly-Supervised Target Atomic Activity Localization}
\label{sec:localization}

\begin{figure*}[t!]
\centering
    \includegraphics[width=18cm]{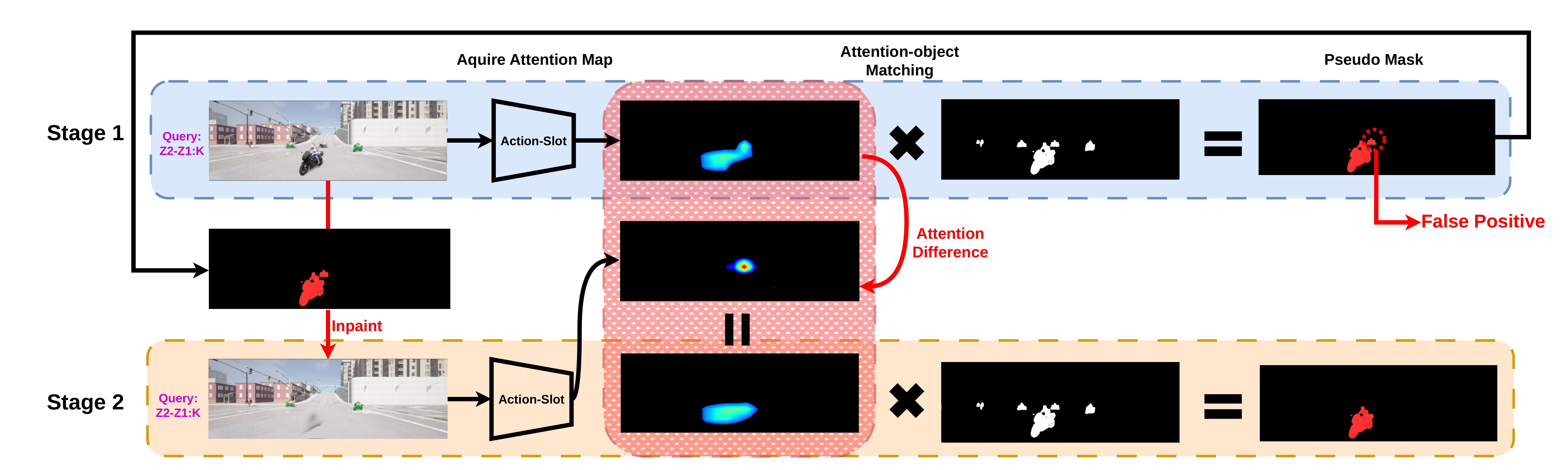}
        \caption{
        \textbf{The Overview of the Proposed Pseudo Mask Selection Framework}. First, we generate the initial pseudo mask by matching the attention map with nearby object masks. To alleviate the false positive resulting from the diffuse or ambiguous attention map, we inpaint the initial pseudo mask region and identify the important regions from the attention difference before and after inpainting, then re-associate with object masks to generate selected pseudo masks.
        }
\label{fig:localization}
\end{figure*}

We extend Action-Slot from video-level recognition to spatio-temporal localization. Our goal is to localize a queried atomic activity using only video-level supervision.
We exploit the attention maps extract from each slot as localization cues and refine them to generate pseudo labels for training.
Specifically, given a queried atomic activity, we extract its corresponding attention map $M_k$ from the associated action slot $S_k$. We perform a two-stage pseudo mask generating procedure -- (1) \textit{Attention-to-Object Matching} and 
(2) \textit{Attention-Difference-Based Pseudo Mask Selection} -- to obtain reliable pseudo masks, as depicted in Figure~\ref{fig:localization}.

%
%
%
%
%

\subsubsection{Attention-to-Object Matching}
\label{sec:matching}
To suppress noise in the raw attention map, we first cluster $M_k$ and align the resulting clusters with object masks. 
Concretely, we apply thresholding to $M_k$ to remove low-confidence attended regions and apply connected component analysis to obtain attention clusters $\{C^k_i\}_{i=1}^{I}$, where the parameter $I$ is the number of connected components.
For the $i^{th}$ cluster $C_i^k$, we first identify the cluster centroid $(c_{x,i}^{k}, c_{y,i}^{k})$ with the object center $(o_{x,j}, o_{y,j})$ of an object mask $Obj_j$ and determine if there is a match via the two criteria:
(1) \textit{Minimum Distance}, which selects the object whose center is closest to the cluster centroid;  
(2) \textit{Gravity-based Distance.} The distance is defined below
\[
\text{gravity}(j) = \frac{\text{area}(Obj_j)}{d^2 + \epsilon},
\]
where $d = \| (c_{x,i}^{k}, c_{y,i}^{k}) - (o_{x,j}, o_{y,j}) \|_2$. 

We use \textit{Gravity-based Distance} for four-wheelers and \textit{Minimum Distance} for two-wheelers and pedestrians. This \textit{mixture matching strategy} aims to mitigate the small-object bias inherent in purely distance-based matching.
Finally, each attention cluster is replaced by the corresponding matched object masks, producing initial pseudo mask $M_k^{\text{init}}$ composed of refined clusters $\{{C'}^k_i\}_{i=1}^{I}$.

%
%

\subsubsection{Attention-Difference-Based Pseudo Mask Selection}
\label{sec:selection}
Although Attention-to-Object Matching provides object-level grounding, the resulting pseudo masks may still include inactive or irrelevant objects due to fragmented or noisy attention responses. 
To further refine the pseudo masks, we introduce an Attention-Difference-Based strategy that estimates the causal effect of each candidate object on the slot attention via counterfactual intervention.
Specifically, we construct a counterfactual video by masking the regions corresponding to the refined clusters $\{{C'}^k_i\}_{i=1}^{I}$ in the initial pseudo mask $M_k^{\mathtt{init}}$. 
We then apply an off-the-shelf video inpainting model, ProPainter~\cite{zhou2023propainter}, to fill in the masked regions, producing an inpainted video $V'$ in which the candidate objects are removed.
Next, we extract the attention map $M^{\mathtt{inpaint}}_k$ from $V'$ and compute the attention difference:
\[
M^{\mathtt{diff}}_k = \max\left(M^{\mathtt{init}}_k - M^{\mathtt{inpaint}}_k, 0\right),
\]
which captures the reduction in attention caused by removing the candidate objects. 
This difference isolates regions that have a genuine causal contribution to a queried atomic activity.
Finally, we apply Attention-to-Object Matching again to $M^{\mathtt{diff}}_k$ to obtain the refined pseudo label $M^{\mathtt{final}}_k$.


 
\subsubsection{Atomic Activity Localization Model}
We adopt a Mask2Former-style localization architecture~\cite{cheng2022masked} to predict the spatio-temporal mask of a queried atomic activity.
The model takes as input the refined attention map $M_k$ corresponding to the queried activity, together with low-level and high-level spatio-temporal features extracted from a shared backbone encoder (e.g., the third and fifth stages of X3D/ResNet), denoted as $F_l$ and $F_h$, respectively.
Notably, the localization module does not directly use slot embeddings as features.
Instead, we perform a cross-attention operation to integrate the slot-specific attention prior with the low-level features $F_l$:
\[
F_{\mathtt{attn}} = \text{Attention}(W_q M_k,\; W_k F_l,\; W_v F_l),
\]
where $W_q$, $W_k$, and $W_v$ are learnable projection matrices.
This operation injects activity-specific priors into the fine-grained feature representation, and that enables the model to focus on regions relevant to the queried atomic activity.
The resulting intermediate feature $F_{\mathtt{attn}}$ is then upsampled and concatenated with the high-level semantic features $F_h$ to form the fused representation $F_{\mathtt{fusion}}$. 
Finally, $F_{\mathtt{fusion}}$ is fed into a lightweight 3D convolutional decoder to predict the spatial-temporal mask $\hat{M}_k$.
The localization model is trained using a combination of binary cross-entropy (BCE) loss and Dice loss~\cite{sudre2017generalised}, defined as $\mathcal{L}_{\mathtt{seg}} = \mathcal{L}_{\mathtt{BCE}} + \mathcal{L}_{\mathtt{Dice}}.$

\section{Experiments}
We investigate the effectiveness of Action-Slot from four perspectives: (1) \textbf{Recognition Performance.} We compare Action-Slot against video-level and object-aware baselines on OATS and TACO, demonstrating the advantage of action-centric representations for multi-label atomic activity recognition. Ablation studies further justify key design choices. 
(2) \textbf{Cross-Domain Generalization.} We evaluate representation transfer by pretraining on TACO and testing on real-world datasets, demonstrating improved generalization and robustness across domains.
(3) \textbf{Structured Decomposition Analysis.} We investigate whether action-centric decomposition is superior to explicit object-level guidance in capturing concurrent atomic activities. 
(4) \textbf{Weakly Supervised Target Atomic Activity Localization Quality.}
We assess the quality of pseudo masks generated at different stages and show that Action-Slot provides stronger localization guidance. Supervising a simple localization model with our selected pseudo masks outperforms existing weakly supervised baselines.

\label{sec:experiments}

\subsection{Datasets}
\subsubsection{\textbf{\textrm{TACO}}} We construct TACO, a balanced and topology-aware benchmark with full atomic activity coverage using the CARLA simulator~\cite{Dosovitskiy17}.

\noindent \textbf{Scenario Collection.}
We adopt three complementary approaches for data collection: auto-pilot, Scenario Runner~\cite{scenario_runner}, and automatic scenario augmentation~\cite{kung2023riskbench}.

\begin{enumerate}[label=(\alph*)]

\item \textbf{Auto-pilot.}
We use the built-in autopilot to control both an ego vehicle and surrounding road users. 
A scenario is automatically recorded when (1) an ego vehicle approaches an intersection and (2) at least one other road user is nearby.
We randomly vary the number and spawn locations of road users to increase diversity. 

\item \textbf{Scenario Augmentation~\cite{kung2023riskbench}.}
We leverage pre-recorded scenarios from RiskBench~\cite{kung2023riskbench}, including both \textit{interactive} (an ego vehicle interacts with risky road users) and \textit{non-interactive} scenarios.
Following the augmentation pipeline of RiskBench, we automatically generate diverse variants by introducing additional road users and altering weather conditions. 
Compared to autopilot-collected data, these scenarios exhibit more human-like maneuvers and riskier interactions. This difference stems from the RiskBench data collection protocol, in which human participants manually control both the ego vehicle and interacting agents.

\item \textbf{Scenario Runner~\cite{scenario_runner}}. The method is developed by CARLA~\cite{Dosovitskiy17} and it enables scripted scenario generation by specifying routes and behaviors for road users.
Unlike auto-pilot and augmentation, it allows us to explicitly manipulate the distribution of queried atomic activities.
A collection starts when an ego vehicle reaches a predefined trigger point and ends after all scripted actions are completed. To enhance diversity, we additionally spawn random road users around an ego vehicle.

\end{enumerate}

For all methods, we randomize weather and lighting, excluding night scenes due to poor visibility.

\noindent \textbf{Sensor Suites.}
We deploy a wide field-of-view camera (120 degrees) to record events taking place on the extreme left and right sides of the ego-vehicle.
We collect images and instance segmentation.
In addition, we collect instance segmentation from Bird's eye view.


\noindent \textbf{Statistics.}
The TACO dataset contains 
5,178 
video clips, of which 1,148 are reserved for testing. In total, we annotate 16,521 video-level atomic activity labels for recognition and 1,939 spatio-temporal annotations for localization. The dataset includes RGB images and instance segmentation maps with an original resolution of 512 $\times$ 1536 pixels. All images are downsampled to 256 $\times$ 768 for model training. Additional dataset statistics are provided in the supplementary material of the CVPR paper~\cite{kungsupplementary}.

\subsubsection{\textbf{\textrm{OATS}}}
The OATS dataset~\cite{Agarwal_2023_ICCV} was collected in San Francisco using an instrumented vehicle. It consists of 1,026 labeled videos and is partitioned into three splits for cross-validation. OATS defines 59 traffic atomic activity categories. Following the original experimental protocol, we train and evaluate on the selected 35 categories. The original image resolution is 1200 × 1920 pixels, and we downsample images to 224 × 224, as in~\cite{Agarwal_2023_ICCV}.

\subsubsection{\textbf{\textrm{nuScenes}}}
We annotate video-level atomic activity labels on the $train\_val$ split of the nuScenes dataset~\cite{nuscenes}, which contains 850 videos. Our annotation protocol produces 426 short clips, each consisting of 16 frames. The annotated nuScenes subset covers 42 atomic activity classes.

 


\subsection{Recognition Performance}

\subsubsection{Comparison with SOTA}\hfill\break
\noindent \textbf{Video-Level Models.}
We implement video-level baselines, including I3D~\cite{carreira2017quo}, X3D~\cite{feichtenhofer2020x3d}, CSN~\cite{tran2019video}, SlowFast~\cite{feichtenhofer2019slowfast}, MViT~\cite{fan2021multiscale}, and VideoMAE~\cite{tong2022videomae}, using the PyTorchVideo library~\cite{fan2021pytorchvideo}.
All models are adapted for multi-label recognition by replacing their original classifiers with a single linear layer that outputs $N_{\mathtt{cl}}$ channels.

\noindent \textbf{Object-Aware Models.}
%
For object-aware baselines, including ORN~\cite{Baradel_2018_ECCV}, ARG~\cite{CVPR2019_ARG}, and OATS~\cite{Agarwal_2023_ICCV}, we first extract object features using an object tracker followed by RoIAlign~\cite{he2017mask}.
%
%
%
We adopt OC-SORT~\cite{maggiolino2023deep} for tracking and set the maximum number of tracklets to 20 following OATS.
To enhance contextual modeling in ORN and ARG~\cite{Baradel_2018_ECCV,CVPR2019_ARG}, we concatenate object features with global video features obtained by applying a $1 \times 1 \times 1$ 3D convolution to X3D features.
Each proposal is classified using a linear layer with $N_{\mathtt{cl}} + 1$ output channels, where the additional class denotes the absence of any action.
We do not include the OATS model~\cite{Agarwal_2023_ICCV} in our comparison due to the unavailability of its official implementation.


\noindent \textbf{Slot-Based Models.}
For slot-based models, including
SAVi~\cite{kipf2022conditional,elsayed2022savi++}, MO~\cite{bao2022discovering}, Slot-VPS~\cite{zhou2022slot}, and our Action-Slot, we use extracted features from backbone encoders such as ResNet~\cite{he2016deep} or X3D~\cite{feichtenhofer2020x3d}.
For X3D, we remove the final projection layer, resulting in slightly smaller model sizes for some slot-based variants, as reported in Table~\ref{table:taco}.
To ensure fair comparison, we re-implement prior slot-based approaches by explicitly assigning slots to atomic activity classes while preserving their original slot initialization strategies.
Specifically, SAVi~\cite{kipf2022conditional} samples slots from a learned distribution at each forward pass, whereas MO~\cite{bao2022discovering} and Slot-VPS~\cite{zhou2022slot} treat slots as learnable queries~\cite{carion2020end}.
Regarding slot update strategies, SAVi and MO perform recurrent temporal updates, while Slot-VPS employs an additional self-attention module~\cite{vaswani2017attention} to update slots across frames.
For all methods, we use the final-frame slot representations as input to the classification head.
%


%


\noindent \textbf{Pretraining and Backbone Choice.}
We first follow the pretraining protocol of OATS~\cite{Agarwal_2023_ICCV} and implement models using a ResNet-50~\cite{he2016deep} backbone pretrained on ImageNet~\cite{deng2009imagenet}.
To further investigate the impact of video pretraining, we additionally train models with video backbones~\cite{carreira2017quo,feichtenhofer2019slowfast,feichtenhofer2020x3d} pretrained on Kinetics-400~\cite{kay2017kinetics} for experiments on both the OATS and TACO datasets. 
Unless otherwise specified, we adopt X3D~\cite{feichtenhofer2020x3d} as the default backbone encoder due to its compact model size.

\subsubsection{Implementation Details}\hfill\break
\noindent \textbf{Background Masks.}
For TACO, background masks are derived from the automatically annotated instance segmentation in CARLA. 
For OATS and nuScenes, we generate background masks using an off-the-shelf DeepLabV3+ model~\cite{chen2018encoder}.

\noindent \textbf{Metrics.}
Following standard practice in multi-label action recognition~\cite{yeung2015every,sigurdsson2016hollywood,gu2018ava,Agarwal_2023_ICCV}, we report mean Average Precision (mAP).

\noindent \textbf{Visualization.}
To qualitatively evaluate whether Action-Slot learns meaningful action-centric representations, we visualize attention maps from the allocated action slots. 
For visualization, we retain pixels with attention scores greater than 0.5 on OATS and 0.2 on TACO. The thresholds are chosen empirically.

\begin{table}[t!]
\centering
\scriptsize
\caption{\textbf{Quantitative results on the OATS dataset.} ``Seq'' denotes the input sequence length. The symbol $\ddag$ denotes the re-implementation of slot-based methods with explicit slot-to-category assignment.
%
\textit{S1}, \textit{S2}, and \textit{S3} correspond to the three predefined splits in the OATS dataset.
}

        \resizebox{0.98\linewidth}!{
        \begin{tabular}
            {@{}l@{\;}c @{\;} c @{\;} | @{\;} c @{\;} c @{\;} c @{\;} c @{\;} | c }
            \toprule
            \multicolumn{1}{l}{Method}& 
            \multicolumn{1}{c}{Backbone}  & 
            \multicolumn{1}{l}{Seq}& 
            \multicolumn{1}{c}{Pretrain}  & 
            \multicolumn{1}{c}{S1}  & 
            \multicolumn{1}{c}{S2}  & 
            \multicolumn{1}{c}{S3}  & 
            \multicolumn{1}{c}{mAP}   
            \\ 
             \midrule
            CSN~\cite{tran2019video,Agarwal_2023_ICCV} & ResNet152 &32 & IG65M & 12.1 & 12.6 & 12.9&12.5
            \\
            TPN~\cite{yang2020temporal,Agarwal_2023_ICCV} & ResNet50 &32 & ImageNet & 11.6& 13.3& 12.9 & 12.6
            \\
            SlowOnly~\cite{feichtenhofer2019slowfast,Agarwal_2023_ICCV} & ResNet50 &32 & ImageNet & 11.2 & 14.7& 12.9 & 13.0
            \\
            SlowFast~\cite{feichtenhofer2019slowfast,Agarwal_2023_ICCV} & ResNet50 &32 & None & 10.8 &15.1 & 14.5 & 13.5
            \\
            I3D (NL)~\cite{wang2018non,Agarwal_2023_ICCV}& ResNet50&32 & ImageNet & 11.9 &15.5 & 14.0 & 13.8
            \\
            I3D~\cite{carreira2017quo,Agarwal_2023_ICCV} &  ResNet50 &32& ImageNet & 11.8 &14.3 &16.8 & 14.3
            \\
            \midrule
             ORN~\cite{Baradel_2018_ECCV,Agarwal_2023_ICCV} & ResNet50 &32& ImageNet & 16.8& 13.4& 18.1 & 16.1
             \\
             ARG~\cite{CVPR2019_ARG,Agarwal_2023_ICCV} & Inceptionv3 &32& ImageNet & 20.2 &21.3 &19.3 & 20.3
             \\
            OATS~\cite{Agarwal_2023_ICCV} & Inceptionv3 &32& ImageNet & \uline{24.3}& \uline{28.6} & \uline{27.2} & \uline{26.7}
            
            \\
            \midrule
             SAVi$\ddag$~\cite{kipf2022conditional,elsayed2022savi++}& ResNet50 &32& ImageNet & 21.1 & 22.4 & 22.5 &22.0
              \\
             MO$\ddag$~\cite{bao2022discovering} &ResNet50 &32& ImageNet &14.3 & 15.6 & 18.2 & 16.0
             \\ 
             Slot-VPS$\ddag$~\cite{zhou2022slot} &ResNet50 &32& ImageNet & 15.7 & 17.8 &17.2 &16.9
              \\
             Action-Slot (Ours) & ResNet50 &32& ImageNet & \textbf{26.6} & \textbf{28.6} & \textbf{30.8} & \textbf{28.6}
              \\
            \hhline{========}
             I3D~\cite{carreira2017quo} & ResNet50 &8& Kinetics-400 & 21.7 & 24.6 &24.4 & 23.6
            \\
             X3D~\cite{feichtenhofer2020x3d} & N/A &16& Kinetics-400 & 30.4	& 33.2 &30.6 & 31.4
              \\
             CSN~\cite{tran2019video} & ResNet101 &32& Kinetics-400 & \uline{43.1} & \uline{47.1} & \uline{44.3} & \uline{44.8}
            \\
             SlowFast~\cite{feichtenhofer2019slowfast} & ResNet50 &16& Kinetics-400 & 36.1	& 36.6 & 34.2 & 35.6
             \\
             \midrule
             ORN~\cite{Baradel_2018_ECCV} & X3D &16& Kinetics-400 & 19.3 & 22.5 & 23.6 & 21.8
             \\
             ARG~\cite{CVPR2019_ARG} & X3D &16& Kinetics-400 & 24.8 & 25.9 & 29.3 & 26.7
             \\
             \midrule
             SAVi$\ddag$~\cite{kipf2022conditional,elsayed2022savi++}& X3D &16& Kinetics-400 & 19.0&22.1& 21.6 &20.9
              \\
             MO$\ddag$~\cite{bao2022discovering} &X3D &16& Kinetics-400 & 25.3 & 25.0 & 24.2& 24.8
             \\
             Slot-VPS$\ddag$~\cite{zhou2022slot} &X3D &16& Kinetics-400 & 24.7	& 24.6 & 25.5 & 24.9
              \\
              Action-Slot (Ours) & X3D &16& Kinetics-400 & \textbf{48.1}& \textbf{47.7}&\textbf{48.8}& \textbf{48.2}
              \\
            \bottomrule
        \end{tabular}
}
\label{table:oats}
\end{table}

\begin{table}[!t]
\centering
\small
\caption{\textbf{Quantitative results on the TACO dataset.} ``Seq'' refers to the input sequence length. \textit{C}, \textit{K}, \textit{P}, \textit{C+}, \textit{K+}, and \textit{P+} denote activity groups involving different road user types.
The symbol $\ddag$ marks the re-implementation of slot-based methods with explicit slot-to-category assignment.
}
\resizebox{1\linewidth}{!}{
        \begin{tabular}
            {@{}l@{\;}c  c  | c @{\;} c@{\;} c @{\;} c @{\;}c @{\;}c @{\;}|c }
            \toprule
            \multicolumn{1}{l}{Method}& 
            \multicolumn{1}{c}{Para. (M)}  & 
            \multicolumn{1}{c}{Seq}  & 
            \multicolumn{1}{c}{C}  & 
            \multicolumn{1}{c}{K}  & 
            \multicolumn{1}{c}{P} &
            \multicolumn{1}{c}{C+} &
            \multicolumn{1}{c}{K+} &
            \multicolumn{1}{c}{P+} &
            \multicolumn{1}{c}{mAP} 
             \\
             \midrule

             I3D~\cite{carreira2017quo} & 27.3 & 8 & 
             27.3 & 19.4 & 30.5 &  34.6&33.6 & 34.8 & 29.7
            \\
             X3D~\cite{feichtenhofer2020x3d} & 3.0 & 16 &
             37.5 & 20.3 & 34.6& \uline{56.3} & 51.5 & 38.8 & 38.3
              \\
             CSN~\cite{tran2019video} &21.4& 32 & 
             \uline{43.5} & \uline{35.5} & \uline{43.0} & 52.5 & \uline{46.1} & \uline{43.4} &\uline{44.0}
            \\
             SlowFast~\cite{feichtenhofer2019slowfast} & 33.7 & 16 & 
             33.9 & 19.7 & 36.7 & 39.9& 42.6 & 41.0 &35.2
                \\
             MViT~\cite{fan2021multiscale} & 36.6 &16 & 
             21.4 & 13.8 & 26.3 & 43.7& 30.0 & 33.8 & 27.9
               \\
             VideoMAE~\cite{tong2022videomae} &  57.9&16 & 
             30.6 & 18.6 & 27.1 &  51.6& 33.0 &37.4 & 33.1
            \\
             \midrule
             ORN~\cite{Baradel_2018_ECCV} & 4.8 & 16 & 25.5 & 15.8 & 24.6 & 31.6 & 19.8 & 13.9 & 22.2 
             \\
            ARG~\cite{CVPR2019_ARG} &12.2& 16 & 27.8 & 15.0 & 27.2 & 35.6 & 13.7 & 20.0 & 23.2
             \\
             \midrule
             SAVi$\ddag$~\cite{kipf2022conditional,elsayed2022savi++}& 2.3 & 16 & 
             19.2 & 13.6 & 27.5 &  23.3& 25.9 & 35.9& 23.3
              \\
             MO$\ddag$~\cite{bao2022discovering} &2.3&16& 
             34.2 & 24.9 & 39.2 & 38.3 & 39.4 & 37.7 & 35.3
             \\
             Slot-VPS$\ddag$~\cite{zhou2022slot} &3.5&16& 
             31.9 & 21.3 & 32.0 & 51.9 & 44.7 & 31.3 & 36.0
              \\
              Action-Slot (Ours) & \textbf{2.3} & 16 & 
              \textbf{48.1} & \textbf{41.2} & \textbf{49.2} & \textbf{70.1} & \textbf{62.6} & \textbf{52.8} &\textbf{54.4}
              \\
            \bottomrule
        \end{tabular}
}
\label{table:taco}
\end{table}

\subsubsection{Main Results}\hfill\break
\label{sec:main}
\noindent \textbf{OATS.}
Table~\ref{table:oats} compares our method with video-level, object-aware, and slot-based approaches.
In the top group, all models use an ImageNet-pretrained ResNet-50 backbone, following the evaluation protocol of OATS~\cite{Agarwal_2023_ICCV}.
In the bottom group, we further compare methods equipped with stronger video backbones pretrained on Kinetics-400~\cite{kay2017kinetics}.
Under both ImageNet and Kinetics pretraining settings, Action-Slot achieves state-of-the-art performance on OATS.
These results validate the advantage of Action-Slot in identifying activity-relevant regions through the proposed structured action-centric representation learning framework,  without relying on object proposals.

\noindent \textbf{TACO.}
We establish a new benchmark on the proposed TACO dataset to evaluate Action-Slot across a broader and more diverse set of atomic activities. 
In the top group of Table~\ref{table:taco}, we additionally include two recent video transformers, MViT~\cite{fan2021multiscale} and VideoMAE~\cite{tong2022videomae}.
These models perform worse than other video-level baselines, which may be attributed to their data-intensive training requirements~\cite{dosovitskiy2021an}.
Object-aware models~\cite{Baradel_2018_ECCV,CVPR2019_ARG,Agarwal_2023_ICCV}, shown in the middle group, achieve suboptimal results, particularly for activities involving groups of road users. 
This suggests that modeling pairwise object relations using MLPs or GCNs~\cite{kipf2016semi} is insufficient for capturing the holistic contextual and motion dynamics required for such activities.
In the bottom group of Table~\ref{table:taco}, we re-implement object-centric slot-based models with allocated slots (denoted by $\ddag$). Action-Slot outperforms all competing methods by a large margin across all activity classes. 
The performance gains can be attributed to our parallel slot updating strategy and the proposed attention guidance mechanism. Notably, Action-Slot excels at activities involving groups of road users, highlighting its ability to model holistic action dynamics. 
Moreover, Action-Slot achieves strong performance while maintaining a compact model size.

\subsubsection{Ablation Study}
\label{sec:ablation}
We conduct ablation studies in Table~\ref{tab:ablation} to evaluate the contributions of the proposed designs.

\noindent \textbf{Non-Allocated vs. Allocated.}
Introducing allocated slots significantly improves performance over the non-allocated variant, yielding a gain of 31.9\% (ID 1 vs. 3). 
These findings underscore the necessity of explicit slot-to-category assignment in tasks with predefined activity taxonomies, where permutation-invariant slot attention may not be well aligned with the atomic activity recognition.

\noindent \textbf{Recurrent vs. Parallel.} 
Updating action slots in parallel across all frames leads to a 17.9\% performance improvement compared to recurrent updating (ID 2 vs. 3). 
This finding indicates that the parallel strategy better captures global spatio-temporal dependencies.

\noindent \textbf{Background Slot and Attention Guidance $L_{\mathtt{bg}}$}.
Simply introducing a non-allocated background slot without mask supervision slightly degrades performance (ID 3 vs. 4), which contrasts with observations in prior slot attention studies~\cite{bao2022discovering,zhou2022slot}. 
We attribute this to potential attention distraction, where the background slot competes with action slots in the absence of explicit supervision.
When supervised with the background mask loss $L_{\mathtt{bg}}$, the model achieves additional performance gains compared to using allocated slots alone (ID 3 vs. 5), demonstrating the effectiveness of attention guidance for stabilizing slot competition.

\begin{table*}[t!]
\begin{minipage}{.4\linewidth}
\centering
\scriptsize
\caption{\textbf{Ablation study of Action-Slot.} The reported results are averaged across the three predefined splits of the OATS dataset.}
\begin{tabular}
            {@{\;} @{\;}c @{\;}|@{\;}c @{\;}|@{\;}c @{\;}|@{\;}c @{\;}|@{\;}c @{\;}|@{\;}c @{\;} | @{\;}c@{\;} }
            \toprule
            \multicolumn{1}{c}{ID}  &
            \multicolumn{1}{c}{Allocated}  &
            \multicolumn{1}{c}{Update}  &
            \multicolumn{1}{c}{BG Slot}  &
            \multicolumn{1}{c}{$L_{\mathtt{bg}}$}  &
            \multicolumn{1}{c}{$L_{\mathtt{neg}}$}  &
            \multicolumn{1}{c}{mAP}
            \\
             \midrule
              1&&parallel&&&&   10.8
              \\
               2&\checkmark&recurrent&&&&  24.8
               \\
              3&\checkmark&parallel&&&&  42.7
              \\
               4&\checkmark&parallel&\checkmark&&&  40.8
              \\
              5&\checkmark&parallel&\checkmark&\checkmark& &  43.6
              \\
              6&\checkmark&parallel&\checkmark& & \checkmark & 43.0
              \\
              7&\checkmark&parallel&\checkmark&\checkmark&\checkmark &  \textbf{48.2}
             \\
            \bottomrule
        \end{tabular}
\label{tab:ablation}
\end{minipage}%
\hspace{0.03\linewidth}
\begin{minipage}{.2\linewidth}

\centering
\scriptsize
\caption{Comparison of backbone architectures for Action-Slot on the TACO dataset.}
\begin{tabular}
            {@{}l@{\;} @{\;} c @{\;}@{\;}}
            \toprule
            \multirow{1}{*}{ \begin{tabular}{@{\;}c@{\;}} \end{tabular}} & 
             \multicolumn{1}{c}{mAP}

             \\
             \midrule
\
             I3D & 29.7
             \\
             Action-Slot &37.6 (+7.9)
             \\
            \midrule
             X3D~\cite{feichtenhofer2020x3d}&  37.8 \\
             Action-Slot & \textbf{54.4} (+16.6)
              \\
             \midrule
             SlowFast~\cite{feichtenhofer2019slowfast}& 35.2
             \\
             Action-Slot & 46.7 (+11.5)
              \\
             
            \bottomrule
        \end{tabular}
        \label{table:backbone}
\end{minipage}
\hspace{0.03\linewidth}
\begin{minipage}{.3\linewidth}
\centering
\scriptsize
\caption{Comparison of Action-Slot and object-level guidance on TACO across different numbers of road users (\textit{N}) per video. \textbf{BG} and \textbf{Neg} represent the background slot and the regularization term in our framework.
}
\begin{tabular}
            {@{}l@{\;} @{\;} c @{\;}@{\;} @{\;}@{\;}c @{\;}@{\;} @{\;} c @{\;} @{\;} c @{\;}  }
            \toprule
            \multirow{1}{*}{ \begin{tabular}{@{\;}c@{\;}} \end{tabular}} & 
             \multirow{1}*{ \begin{tabular}{@{}c@{}} $N$ $\leq$ 5 \end{tabular}} & 
             \multirow{1}*{ \begin{tabular}{@{}c@{}}  5 $<$ $N$ $\leq$ 15 \end{tabular}} & 
             \multirow{1}*{ \begin{tabular}{@{}c@{}} $N$ $>$ 15 \end{tabular}} 
             &
             
             \\
             \midrule
             object & 49.4 & 47.7 & 43.5
             \\
             BG+Neg & \textbf{55.2} & \textbf{50.9} & \textbf{46.3}
             \\
            \bottomrule
        \end{tabular}
        
        \label{table:object_guidance}
\end{minipage}
\end{table*}

\noindent \textbf{Action Slots Regularization $L_{\mathtt{neg}}$.}
We regularize action slots with the $L_{\mathtt{neg}}$ term to discourage action slots allocated to negative classes from attending any regions in spatio-temporal features.
%
%
This design enhances the likelihood of other action slots allocated to positive classes identifying the region of interest more effectively. 
We demonstrate that the inclusion of the regularization term $L_{\mathtt{neg}}$ further improves the one only using the allocated slot (ID 3 vs. 6).

The final configuration (ID 7 in Table~\ref{tab:ablation}) consistently achieves the best performance on OATS, providing comprehensive validation of the proposed Action-Slot design.

\noindent \textbf{Action-Slot with Different Backbones.}
Table~\ref{table:backbone} reports the performance of Action-Slot using different backbone encoders.
%
We choose I3D~\cite{carreira2017quo}, X3D~\cite{feichtenhofer2020x3d}, and SlowFast~\cite{feichtenhofer2019slowfast}  
for experiments.
Action-Slot significantly improves all 3D CNN methods, showing great generalization ability.
%
Among them, X3D achieves the most improvement.
We hypothesize that this is because X3D's features retain the original temporal dimension, affording Action-Slot greater spatio-temporal information to decompose atomic activities.
%
In contrast, other backbone encoders downsample the temporal dimension; for instance, SlowFast takes a video sequence of length 16 as input and reduces the temporal dimension to 4.
%
%

\begin{table*}[!t]
\begin{center}
\begin{tabular}{cc@{\;}c@{\;}c@{\;}c@{\;}c@{\;}c}
\vspace{.2cm}
\raisebox{-.4\height}{\rotatebox{90}{\makecell{\scriptsize Action-Slot}}}
\hspace{-4mm}
& \adjustimage{height=2.13cm,valign=m}{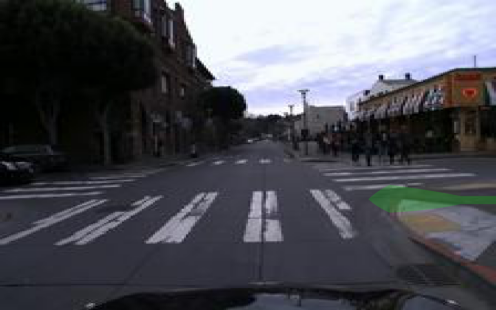}
& \adjustimage{height=2.13cm,valign=m}{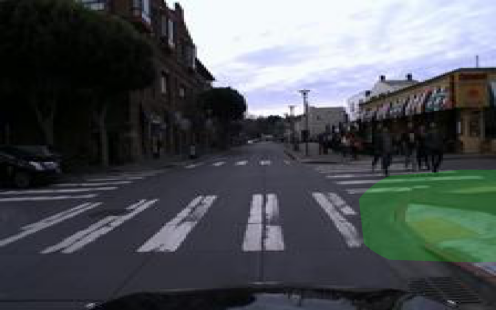}
& \adjustimage{height=2.13cm,valign=m}{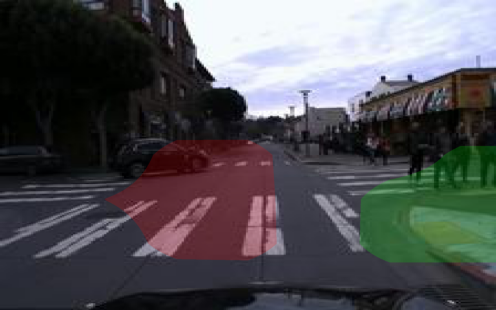}
& \adjustimage{height=2.13cm,valign=m}{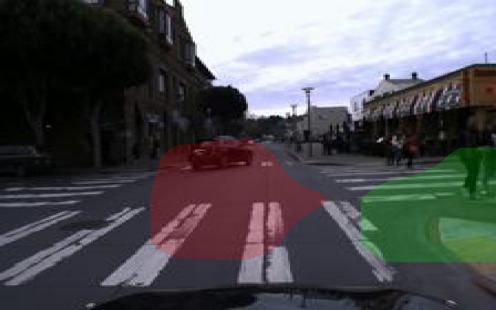}
& \adjustimage{height=2.13cm,valign=m}{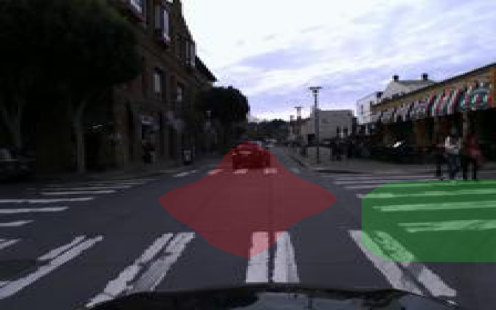}
\hspace{.04cm}
\\
\vspace{-.2cm}

\raisebox{-.4\height}{\rotatebox{90}{\makecell{\scriptsize MO$\ddag$~\cite{bao2022discovering}}}}
\hspace{-4mm}
& \adjustimage{height=2.13cm,valign=m}{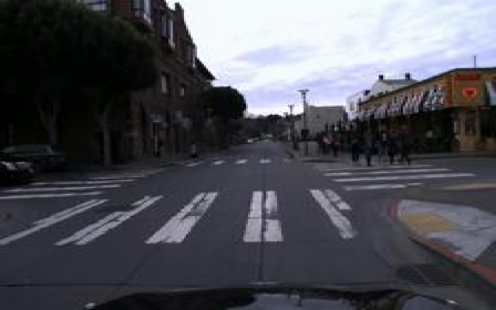}
& \adjustimage{height=2.13cm,valign=m}{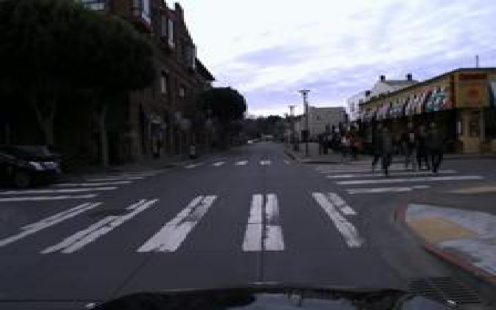}
& \adjustimage{height=2.13cm,valign=m}{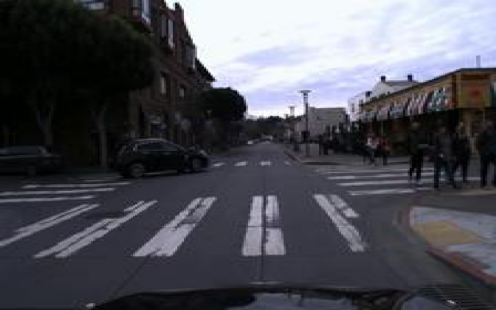}
& \adjustimage{height=2.13cm,valign=m}{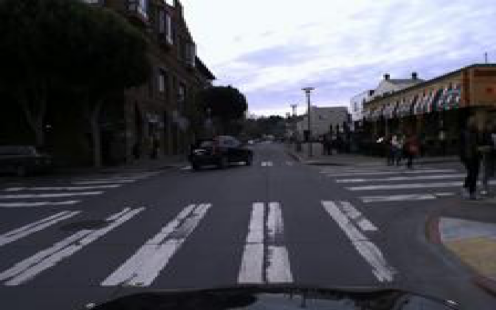}
& \adjustimage{height=2.13cm,valign=m}{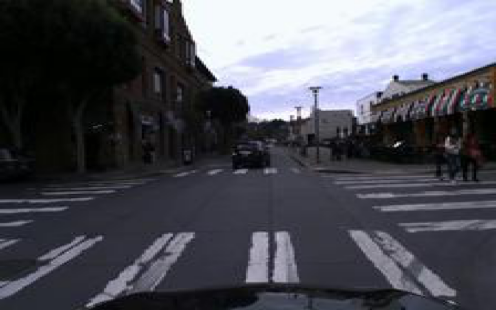}
\hspace{.04cm}

\end{tabular}
\end{center}
\captionof{figure}{
\textbf{Visualization of attention maps learned on OATS.} Colored masks depict the attention of the activity slots \textcolor{red}{Z4-Z3:C} and \textcolor{green}{C2-C1:P+}. 
Note that, while MO successfully predicts the activity \textcolor{green}{C2-C1:P+}, the corresponding attention scores are very low.}
\label{fig:attention_oats}
\end{table*}

\begin{table*}[!t]
\begin{center}
\begin{tabular}{cc@{\;}c@{\;}c@{\;}c@{\;}c@{\;}c}

\vspace{.2cm}
\raisebox{-.4\height}{\rotatebox{90}{\makecell{\scriptsize Action-Slot}}}
\hspace{-4mm}
& \adjustimage{height=1.14cm,valign=m}{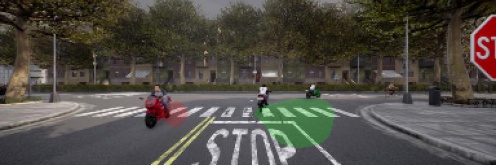}
& \adjustimage{height=1.14cm,valign=m}{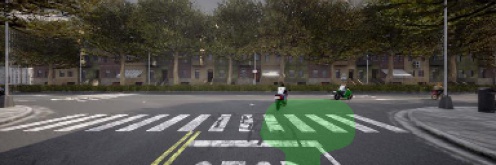}
& \adjustimage{height=1.14cm,valign=m}{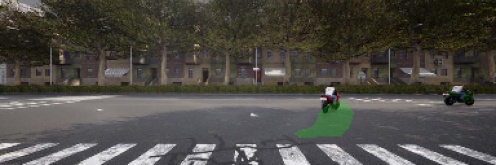}
& \adjustimage{height=1.14cm,valign=m}{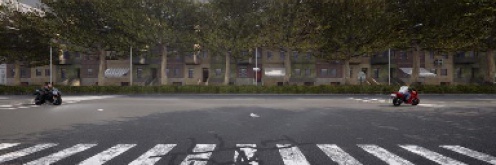}
& \adjustimage{height=1.14cm,valign=m}{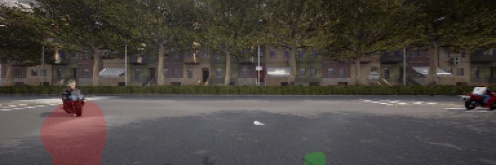}
\\
\hspace{.04cm}

\vspace{-.2cm}

\hspace{-6mm} 
\raisebox{-.4\height}{\rotatebox{90}{\makecell{\scriptsize MO$\ddag$~\cite{bao2022discovering}}}}
\hspace{-8mm}
& \adjustimage{height=1.14cm,valign=m}{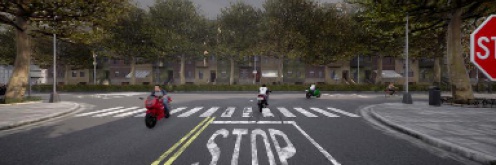}
& \adjustimage{height=1.14cm,valign=m}{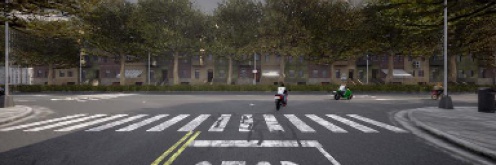}
& \adjustimage{height=1.14cm,valign=m}{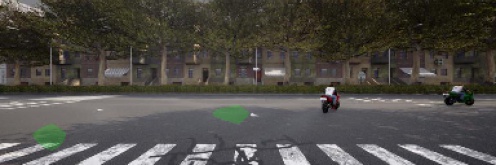}
& \adjustimage{height=1.14cm,valign=m}{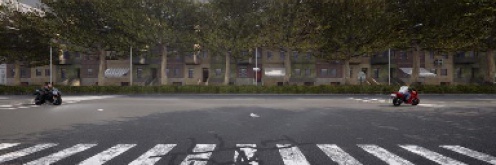}
& \adjustimage{height=1.14cm,valign=m}{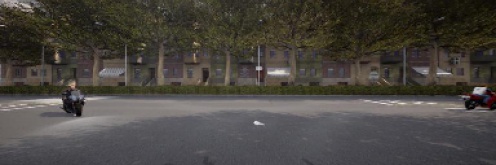}

\end{tabular}
\end{center}
\captionof{figure}{
\textbf{Visualization of attention maps learned on TACO.} Colored masks are from the allocated slots of activity \textcolor{red}{Z4-Z1:K} and \textcolor{green}{Z1-Z2:K+}. Note that \textcolor{red}{Z4-Z1:K} appears twice in the 1$^{\textrm{st}}$ and 5$^{\textrm{th}}$ frames independently. 
}
\label{fig:attention_taco}
\end{table*}


             

\subsection{Cross-Domain Generalization}
To assess generalization, we perform transfer learning from TACO to real-world benchmarks, including OATS~\cite{Agarwal_2023_ICCV} and a newly annotated nuScenes dataset~\cite{nuscenes}. We compare pretraining on Kinetics, OATS, and TACO (Table~\ref{table:transfer}), followed by fine-tuning on the respective target datasets.
%

\noindent \textbf{Pretraining on TACO.}
Due to its large scale and relatively balanced activity distribution, TACO pretraining consistently improves performance across all models on both real-world datasets.

\noindent \textbf{Sim-to-Real Transferability.}
Although TACO is collected in a synthetic environment, it demonstrates stronger cross-domain generalization than OATS. 
As shown in Table~\ref{table:transfer}, OATS pretraining yields marginal improvements and even degrades performance for X3D, whereas TACO pretraining results in substantial gains across all three categories of models. 
These findings validate the effectiveness of large-scale, low-cost synthetic data in enhancing sim-to-real transfer for atomic activity recognition.

\begin{table}[h]

\centering
\scriptsize
\caption{
Comparisons of pretrained representations from OATS
and TACO. We perform transfer learning on the real-world OATS and nuScenes dataset. 
We report the average mAP of the three splits in the OATS dataset.
}
\vspace{-2mm}
\resizebox{0.9\linewidth}{!}{
\begin{tabular}
            {@{}l  c c   c  c c }
            \toprule
            \multirow{2}{*}{ \begin{tabular}{@{\;}c@{\;}} \\\end{tabular}} & 

            \multicolumn{2}{c}{OATS}  & 
             \multicolumn{3}{c}{nuScenes}  

             \\
             \cmidrule(lr){2-3} \cmidrule(lr){4-6}
            &
             \begin{tabular}{c@{\;}}  Kinetics  \end{tabular} & 
             \begin{tabular}{c}  +TACO   \end{tabular} & 
            \begin{tabular}{c@{\;}} Kinetics   \end{tabular} & 
            \begin{tabular}{c@{\;}} +OATS   \end{tabular} & 
             \begin{tabular}{c@{\;}} +TACO  \end{tabular} 
            
              \\
            \midrule
            X3D~\cite{feichtenhofer2020x3d} & 31.4& 34.9 (+3.5) & 19.8& 18.9 (-0.9) & 27.8 (+8.0)
            \\
            ARG~\cite{CVPR2019_ARG} & 26.7 & 31.3 (+4.6) & 12.2 &12.7 (+0.5) & 17.0 (+4.8)
            \\
             Action-Slot & 48.2 & \textbf{59.5} (+11.3) & 23.6& 23.6 (+0.0)& \textbf{32.3} (+8.7)
             \\
            \bottomrule
        \end{tabular}
        
        \label{table:transfer}
        }
\end{table}

\subsection{Structured Decomposition Analysis}
\label{sec:Structured-Decomposition-Analysis}
To verify that Action-Slot truly performs activity-centric decomposition rather than object grouping, we compare against object-level guidance and visualize the differences.

\noindent \textbf{Action-Slot vs. Object-level Guidance.}
We conduct experiments on the TACO dataset to evaluate the effect of varying numbers of road users per video under two attention guidance mechanisms within the Action-Slot framework
%
%
For object-level guidance, we use instance segmentation collected in TACO to obtain masks for vehicles and pedestrians.
Hungarian matcher is used to assign object masks to slots, following MO~\cite{bao2022discovering}.
%
%
%
As shown in Table~\ref{table:object_guidance}, Action-Slot with strong ground-truth object-level supervision performs worse than our proposed attention guidance mechanism.
%
This result suggests that relying solely on object-level cues is insufficient for our task, as not all objects participate in the same atomic activity at any given time.
%

\noindent \textbf{Qualitative Results.}
We visualize the attention maps learned using the re-implemented MO~\cite{bao2022discovering} and Action-Slot on the OATS and TACO datasets in Figure~\ref{fig:attention_oats} and Figure~\ref{fig:attention_taco}, respectively.
%
%
%
Action-Slot demonstrates the capability to 
decompose multiple atomic activities from intricate scenes on both datasets. 
In Figure~\ref{fig:attention_oats}, Action-Slot localizes the group activity \textbf{C2-C1:P+}, while MO~\cite{bao2022discovering} fails to attend any relevant regions. 
Moreover, Action-Slot recognizes two independent \textbf{Z4-Z1:K} activities, instead of predicting them as a group activity, i.e., \textbf{Z4-Z1:K+} in Figure~\ref{fig:attention_taco}.
The evidence justifies the capability of Action-Slot to decompose structured activities. 
%
More interestingly, 
%
Action-Slot
attends to relevant regions when road users initiate an action and cease attention once the action is completed, as seen in the case of the bicyclists in the fourth frame of Figure~\ref{fig:attention_taco}.


%
%
%


\subsection{Weakly-Supervised Target Atomic Activity Localization Quality}
%
We investigate spatio-temporal localization to examine two central questions:
(1) Do the action-centric representations learned by Action-Slot inherently capture spatial-temporal grounding signals suitable for pseudo-mask generation? 
(2) How to enhance the coarse spatial-temporal cues and enable reliable localization?

\begin{table*}[t!]
\centering
\caption{Evaluation of pseudo masks across different generation stages. The source action recognition models are trained exclusively on recognition task.}
\begin{tabular}{llccccccc}
\toprule
\textbf{Stage} & \textbf{Attention map Source} & \textbf{mIoU} & \textbf{oIoU} & \textbf{tIoU} & \textbf{mAP@tIoU} & \textbf{Prec.} & \textbf{Recall} & \textbf{F1} \\
\midrule
\multirow{4}{*}{Raw Attention/Grad-CAM} 
    & X3D~\cite{feichtenhofer2020x3d}    & 3.6& 3.4& 0.9& 0& 6.1& 18.8& 6.5\\
    & CSN~\cite{tran2019video}           & 2.0& 2.5& 0.5& 0& 2.9& 9.3& 3.3\\
    & Slot-VPS~\cite{zhou2022slot}       & 1.8  & 2.9  & 0.6  & 0    & 4.8  & 5.3  & 3.2  \\
    & Action-Slot                        & 4.6  & 5.6  & 1.7  & 0    & 9.1  & 15.7 & 8.0  \\
\midrule
\multirow{4}{*}{Attention-Object Matching} 
    & X3D~\cite{feichtenhofer2020x3d}    & 16.1& 21.0& 15.7& 3.4& 22.5& 31.4& 22.3\\
    & CSN~\cite{tran2019video}           & 17.7& 23.2& 17.3& 3.8& 23.4& 36.0& 24.2\\
    & Slot-VPS~\cite{zhou2022slot}       & 14.6& 21.3& 14.6& 3.5& 30.1& 19.6& 20.1\\
    & Action-Slot   & 23.0& 24.0& 25.4& 9.1& 41.1& 31.9& 30.7\\
\midrule
\multirow{4}{*}{Pseudo Mask Selection}
    & X3D~\cite{feichtenhofer2020x3d}    & 16.9& 20.9& 16.5& 4.1& 23.8& 32.1& 23.0\\
    & CSN~\cite{tran2019video}           & 17.8& 23.3& 17.4& 3.8& 23.3& 36.1& 24.2\\
    & Slot-VPS~\cite{zhou2022slot}       & 18.0& 23.1& 17.8& 3.5& 34.2& 25.5& 24.9\\
    & Action-Slot   & \textbf{26.8}& \textbf{29.8}& \textbf{28.8}& \textbf{9.6} & \textbf{47.0}& \textbf{36.9}& \textbf{35.9}\\
\bottomrule
\end{tabular}
\label{tab:pseudo-mask-quality-predicted}
\end{table*}
\subsubsection{Implementation Details}
%

To bridge the gap between video-level recognition and pixel-level localization, we leverage foreground object masks generated by Grounded-SAM-2~\cite{ravi2024sam2segmentimages} as spatial proposals used in the matching process described in Sec.~\ref{sec:matching}, yielding high-quality pseudo-masks for supervision. 
We then train the localization model with these pseudo-masks with the segmentation loss $\mathcal{L}_{\mathtt{seg}}$ to fine-tune the entire Action-Slot framework.

For the ablation study on attention-object matching in Sec.~\ref{sec:loc_ablation}, we implement the baseline pseudo-mask generation algorithms proposed by Chen \textit{et al.}~\cite{chen2020learning}, which converting attention maps into initial binary masks $M_{\mathtt{init}}$ using Otsu's thresholding~\cite{otsu1975threshold}. The baseline originally refine masks by merging superpixels~\cite{achanta2012slic}; however, we replace the superpixels with the predicted object masks for a fair comparison and merge those that yield an IoU $\geq0.1$ with $M_{\mathtt{init}}$.



\subsubsection{Baselines}
To demonstrate the spatial-temporal transferability of Action-Slot, we compare the pseudo masks generated by other recognition models, i.e., X3D, CSN, and Slot-VPS.
For X3D and CSN, we compute Grad-CAMs~\cite{selvaraju2017grad} from the last layer before the projection head and use them as pseudo masks. The Grad-CAM outputs are trilinearly interpolated to match the spatial-temporal resolution of Action-Slot’s attention maps. For Slot-VPS, we directly adopt its attention maps as pseudo masks.

To compare the proposed weakly supervised localization model with the existing approaches, we re-implement the multiple instance learning (MIL) method proposed by Chen \textit{et al.}~\cite{chen2023weakly}.
Specifically, we replace the RPN-generated object proposals~\cite{ren2015faster} with object masks generated by Grounded SAM 2~\cite{ravi2024sam2segmentimages} and omit the unsupervised tracking stage.
We also re-implement VSCR~\cite{duan2024mining} by incorporating its CAM-aware contrastive loss while replacing the original CAM map with our generated initial pseudo mask.


\begin{table*}[t!]
\centering
\caption{Performance of localization models trained on different pseudo masks. Compared to existing weakly-supervised action localization baselines.}
\begin{tabular}{lccccccc}
\toprule
\textbf{Method} & \textbf{mIoU} & \textbf{oIoU} & \textbf{tIoU} & \textbf{mAP@tIoU} & \textbf{Precision} & \textbf{Recall} & \textbf{F1} \\
\midrule
Chen \textit{et al.}~\cite{chen2023weakly} & 15.0& 16.1& 14.2& 2.5& 16.8& 35.8& 20.5\\
VSCR~\cite{duan2024mining} & 22.8& 25.6& 20.2& 5.5& 35.4& 44.9& 32.2\\
Ours w/o selection  & 22.7& 25.7& 20.3& 5.7& 34.6& \textbf{45.3}& 31.2\\
Ours & \textbf{24.7}& \textbf{28.1}& \textbf{23.5}& \textbf{6.8}& \textbf{38.9}& 44.0& \textbf{34.8}\\
\bottomrule
\end{tabular}
\label{tab:localization-performance}
\end{table*}

\begin{table}[t!]
\centering
\caption{
False Positive (FP) rates (\%) for absent-action queries across different actor types. A sample is counted as an FP if the number of activated pixels exceeds an actor-specific threshold.
}
\begin{tabular}{lccc}
\toprule
\textbf{Method} & \textbf{FP(C/C+)} & \textbf{FP(K/K+)} & \textbf{FP(P/P+)} \\
\midrule
Ours w/o Selection& 53.4& 59.2& 76.8\\
Ours w/ Selection& \textbf{25.2}& \textbf{32.8}& \textbf{49.4}\\
\bottomrule
\end{tabular}
\label{tab:fp-rate}
\end{table}

\subsubsection{Metrics} 
We evaluate localization performance using mean IoU (mIoU), overall IoU (oIoU), and temporal IoU (tIoU).
The mIoU computes the average IoU across actions, while oIoU aggregates all predicted and ground-truth pixels over the dataset before computing IoU, emphasizing larger samples.
Since our model outputs spatio-temporal masks rather than per-frame confidence scores, we redefine tIoU using frame-wise spatial IoU as a proxy for confidence. 
Frames with spatial IoU greater than 0.5 are considered true positives, and tIoU is computed as the ratio of true positive frames to the total number of frames containing either predicted or ground-truth pixels.
In addition, we report pixel-level precision, recall, and F1 score to provide a more comprehensive evaluation of mask quality.

\subsubsection{Pseudo Mask Quality}
\label{sec:pseudo}
We evaluate the quality of pseudo masks generated by models trained exclusively on the recognition task. As shown in Table~\ref{tab:pseudo-mask-quality-predicted},
masks derived from raw attention maps and Grad-CAM~\cite{selvaraju2017grad} are highly inaccurate, yielding coarse localization.
After applying the attention-to-object matching introduced in Sec.~\ref{sec:matching}, all methods exhibit significant improvements across all evaluation metrics.
This indicates that existing action recognition models fail to offer fine object localization cues.
To further refine these masks leverages the important regions identified through attention differences (described in Sec.~\ref{sec:selection}), leading to consistent gains across all metrics.
These results validate the effectiveness of our selection strategy in isolating the most relevant spatio-temporal regions associated with a specific action.
Notably, at every stage, pseudo masks generated by Action-Slot consistently outperform those from other models, suggesting that its action-centric representation provides crucial action semantics enabling strong spatio-temporal localization.

\begin{table*}[t!]
\centering
\begin{tabular}{c@{\;}c@{\;}c@{\;}c@{\;}c@{\;}c@{\;}c}
\vspace{1mm}
\raisebox{-.3\height}{\rotatebox{90}{\makecell{\scriptsize Frame}}} &
\adjustimage{height=1.12cm,valign=m}{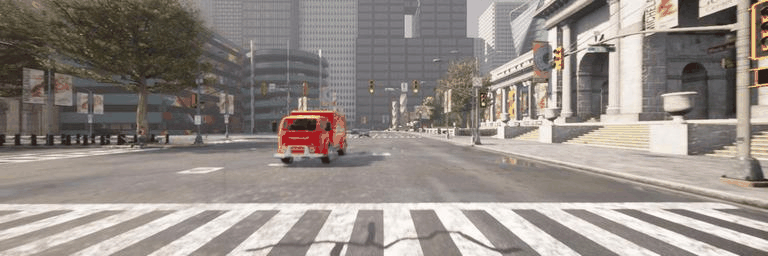} &
\adjustimage{height=1.12cm,valign=m}{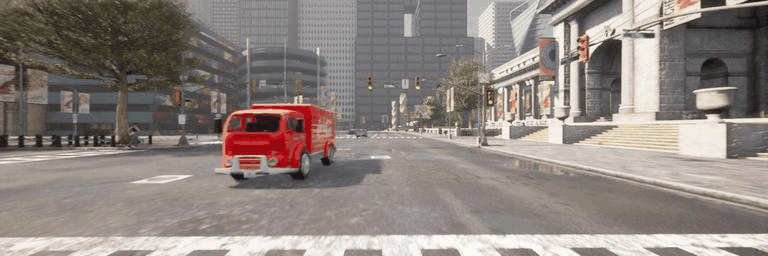} &
\adjustimage{height=1.12cm,valign=m}{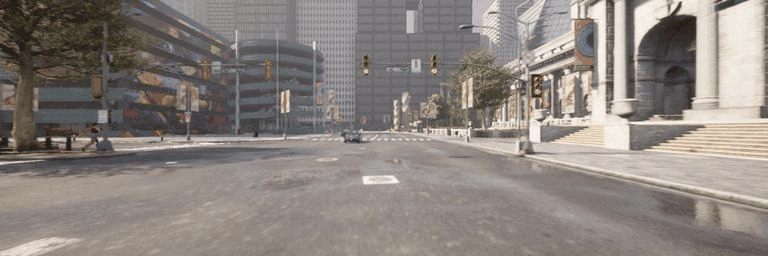} &
\adjustimage{height=1.12cm,valign=m}{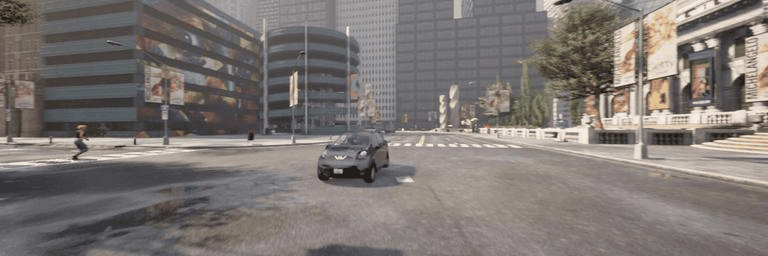} &
\adjustimage{height=1.12cm,valign=m}{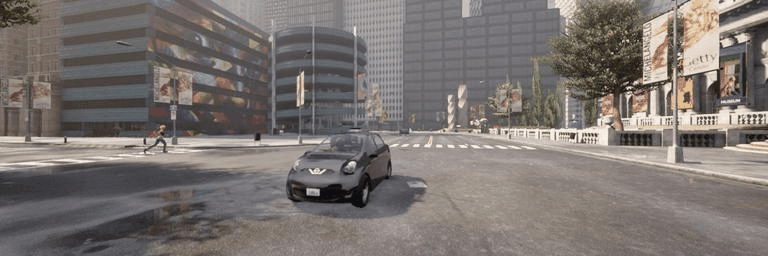} \\[1mm]
\vspace{1mm}

\raisebox{-.3\height}{\rotatebox{90}{\makecell{\scriptsize Z3-Z1:C+}}} &
\adjustimage{height=1.12cm,valign=m}{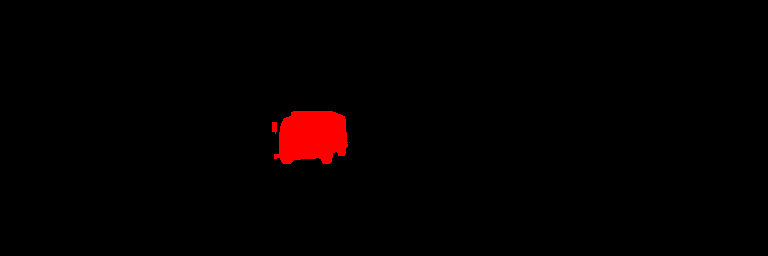} &
\adjustimage{height=1.12cm,valign=m}{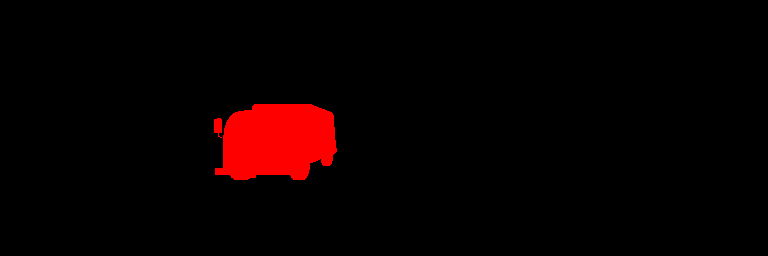} &
\adjustimage{height=1.12cm,valign=m}{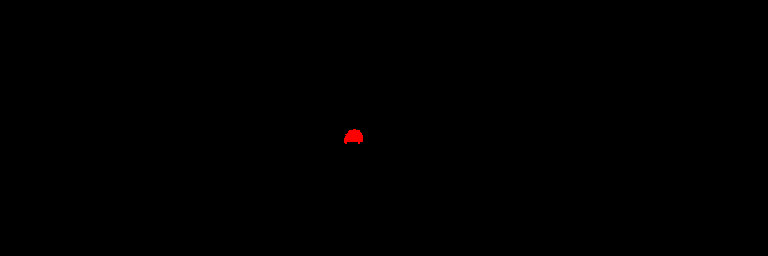} &
\adjustimage{height=1.12cm,valign=m}{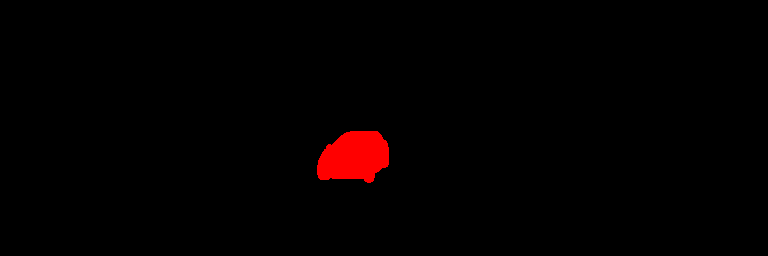} &
\adjustimage{height=1.12cm,valign=m}{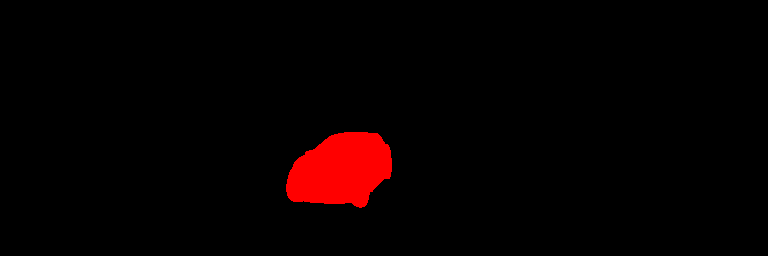} \\[1mm]
\vspace{1mm}

\raisebox{-.4\height}{\rotatebox{90}{\makecell{\scriptsize Ours w/o \\ Selection}}} &
\adjustimage{height=1.12cm,valign=m}{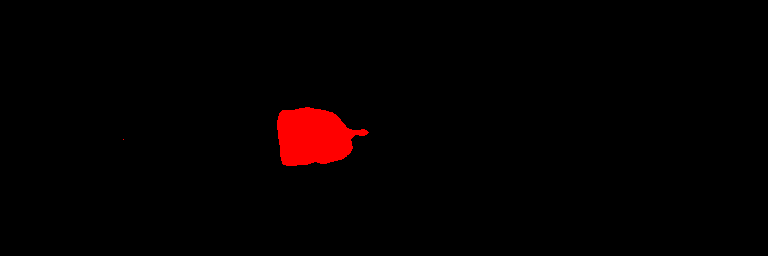} &
\adjustimage{height=1.12cm,valign=m}{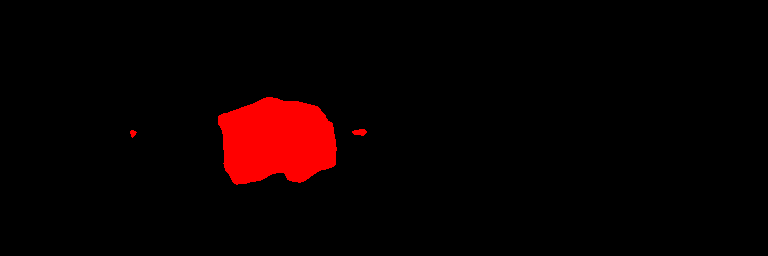} &
\adjustimage{height=1.12cm,valign=m}{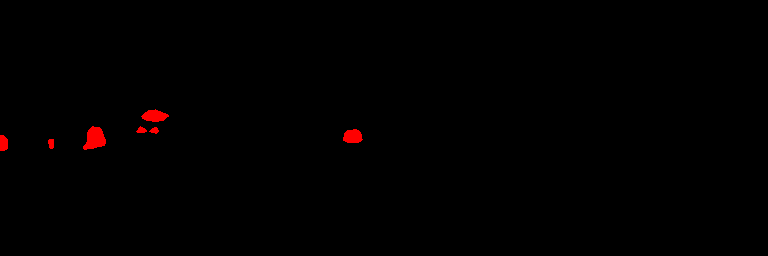} &
\adjustimage{height=1.12cm,valign=m}{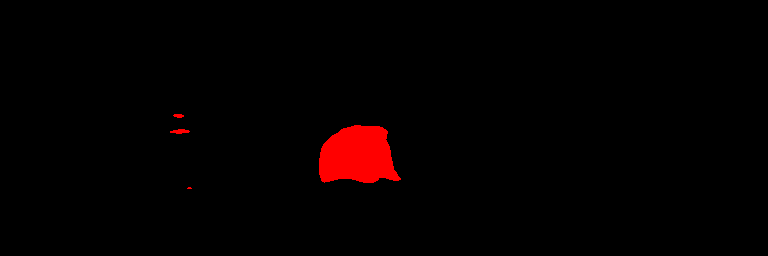} &
\adjustimage{height=1.12cm,valign=m}{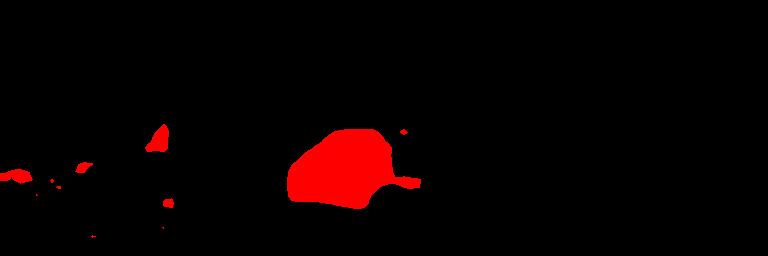} \\[1mm]
\vspace{1mm}

\raisebox{-.3\height}{\rotatebox{90}{\makecell{\scriptsize Ours}}} &
\adjustimage{height=1.12cm,valign=m}{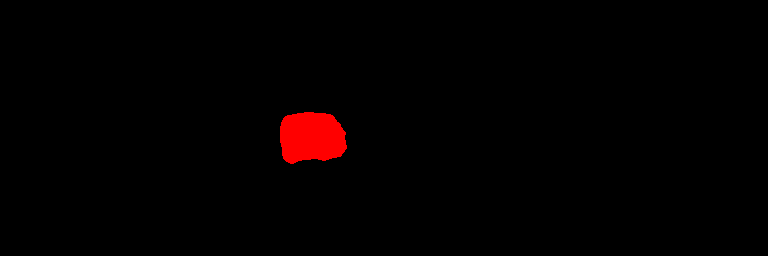} &
\adjustimage{height=1.12cm,valign=m}{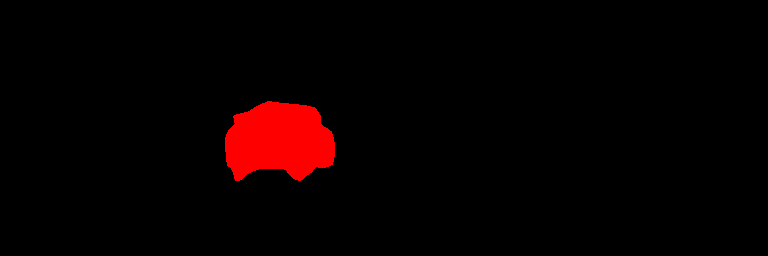} &
\adjustimage{height=1.12cm,valign=m}{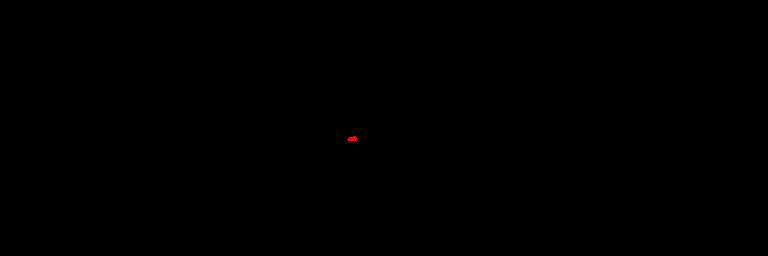} &
\adjustimage{height=1.12cm,valign=m}{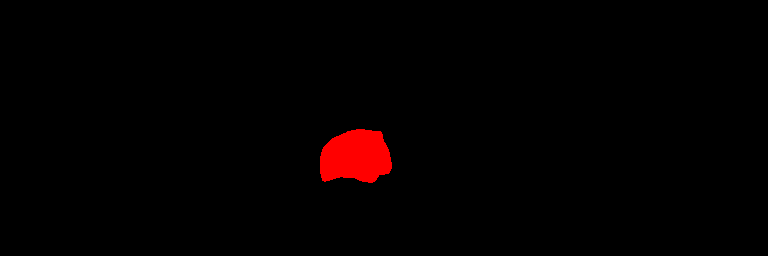} &
\adjustimage{height=1.12cm,valign=m}{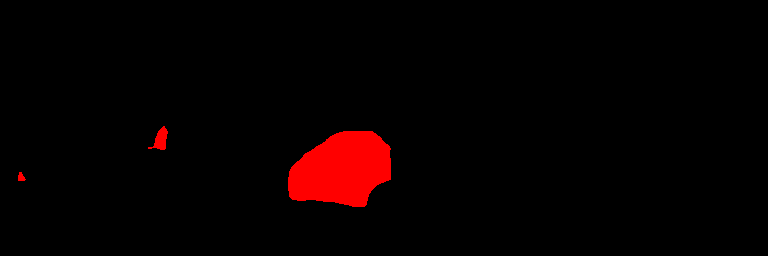} \\

\vspace{1mm}
\raisebox{-.3\height}{\rotatebox{90}{\makecell{\scriptsize Frame}}} &
\adjustimage{height=1.12cm,valign=m}{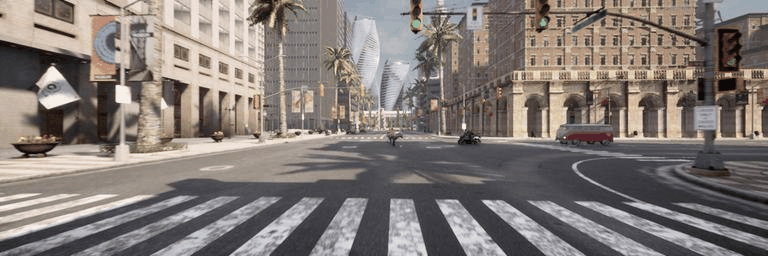} &
\adjustimage{height=1.12cm,valign=m}{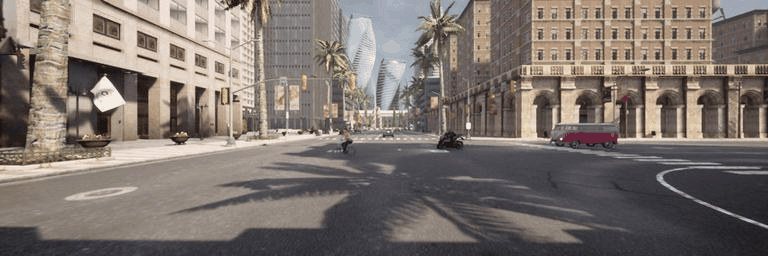} &
\adjustimage{height=1.12cm,valign=m}{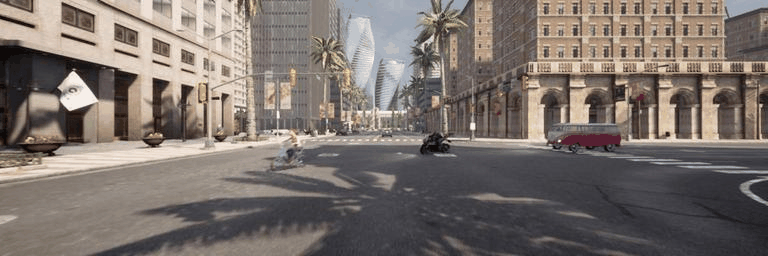} &
\adjustimage{height=1.12cm,valign=m}{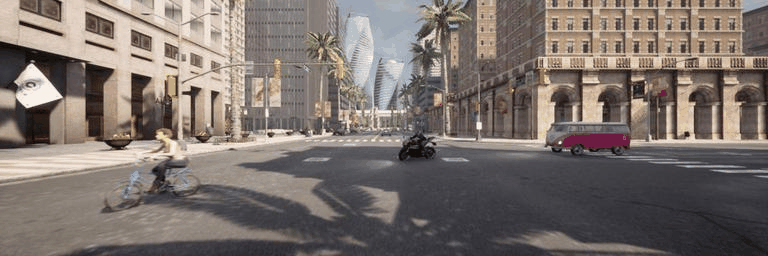} &
\adjustimage{height=1.12cm,valign=m}{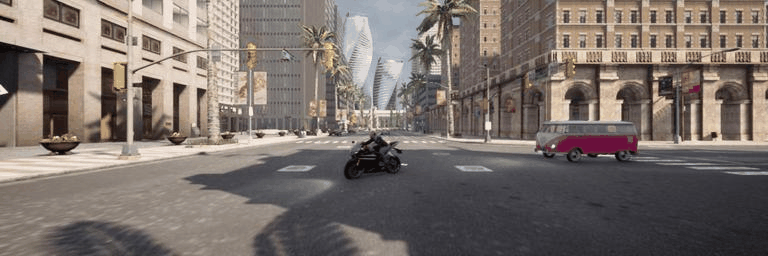} \\[1mm]
\vspace{1mm}

\raisebox{-.3\height}{\rotatebox{90}{\makecell{\scriptsize Z2-Z1:K+}}} &
\adjustimage{height=1.12cm,valign=m}{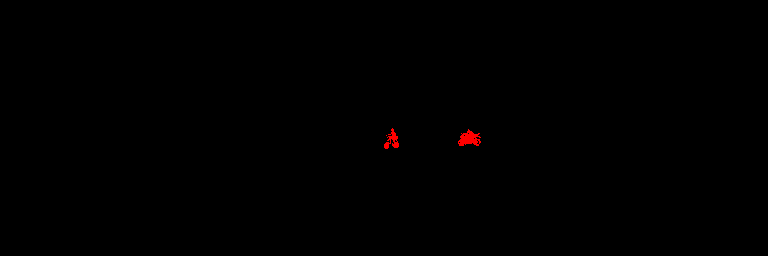} &
\adjustimage{height=1.12cm,valign=m}{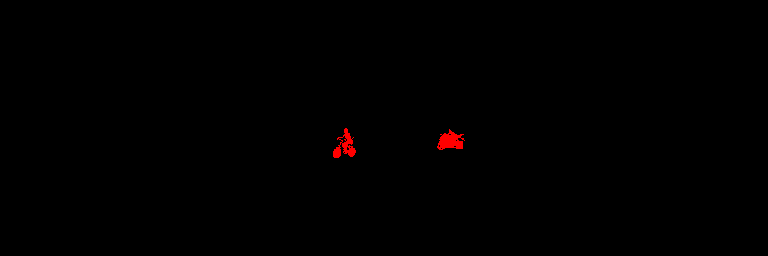} &
\adjustimage{height=1.12cm,valign=m}{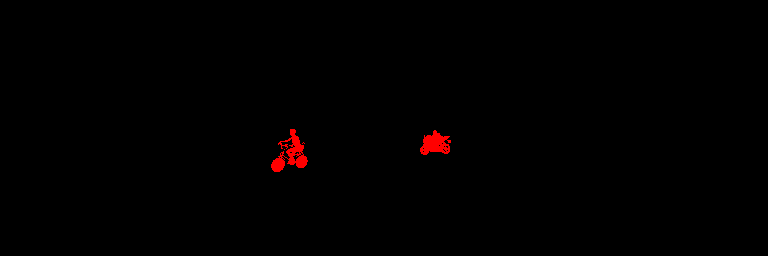} &
\adjustimage{height=1.12cm,valign=m}{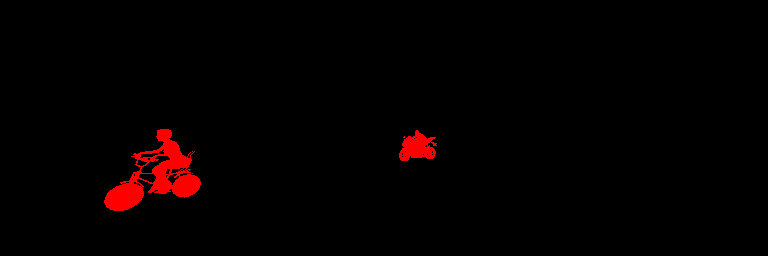} &
\adjustimage{height=1.12cm,valign=m}{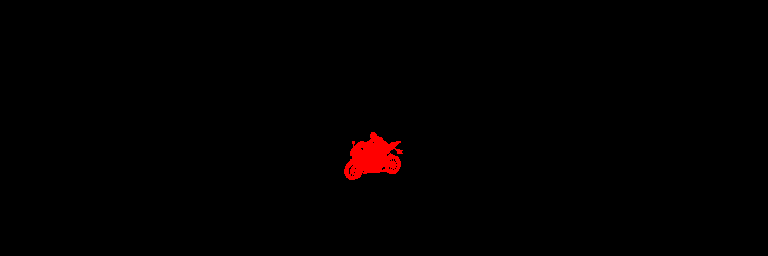} \\[1mm]
\vspace{1mm}

\raisebox{-.4\height}{\rotatebox{90}{\makecell{\scriptsize Ours w/o \\ selection}}} &
\adjustimage{height=1.12cm,valign=m}{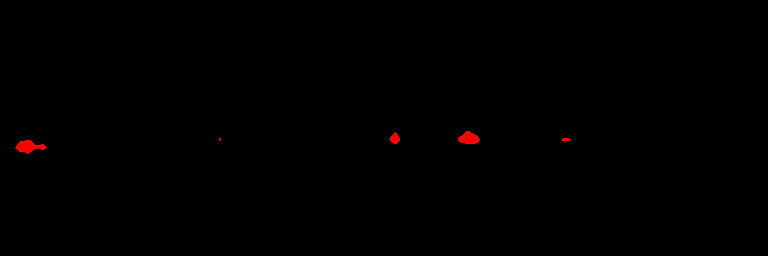} &
\adjustimage{height=1.12cm,valign=m}{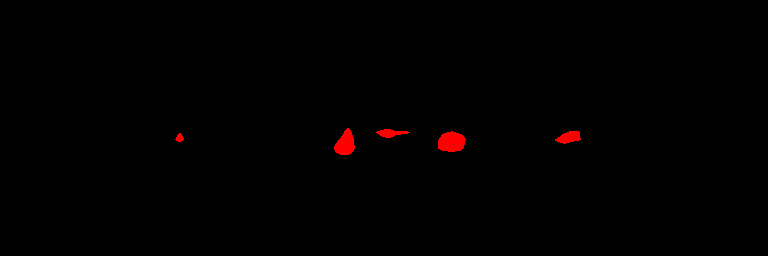} &
\adjustimage{height=1.12cm,valign=m}{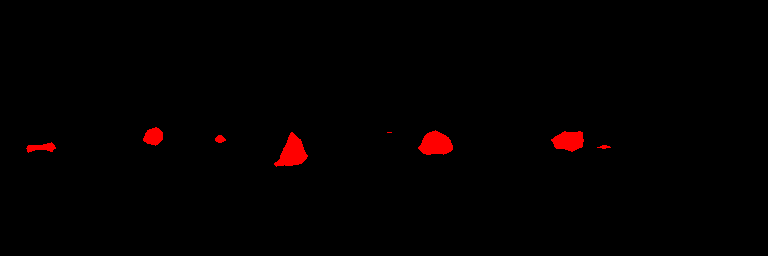} &
\adjustimage{height=1.12cm,valign=m}{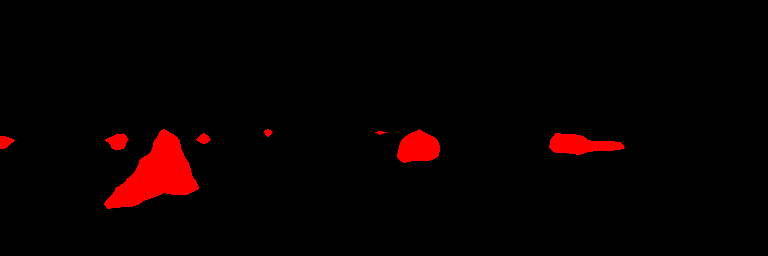} &
\adjustimage{height=1.12cm,valign=m}{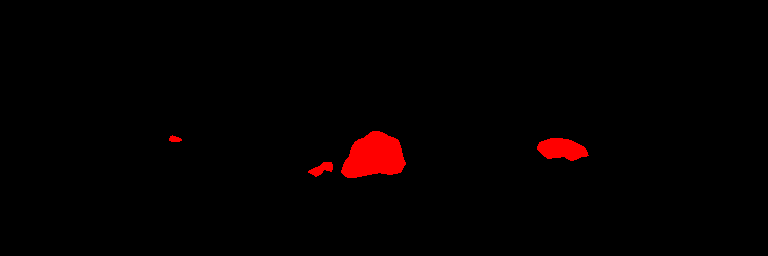} \\[1mm]
\vspace{1mm}

\raisebox{-.3\height}{\rotatebox{90}{\makecell{\scriptsize Ours}}} &
\adjustimage{height=1.12cm,valign=m}{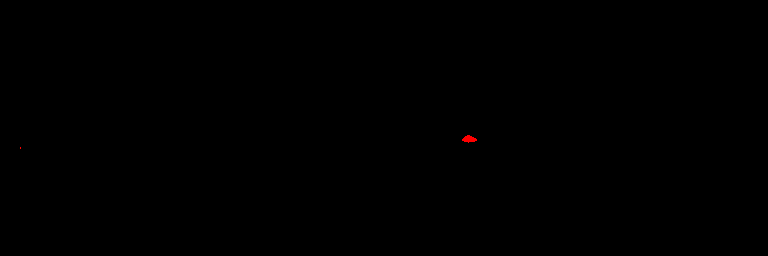} &
\adjustimage{height=1.12cm,valign=m}{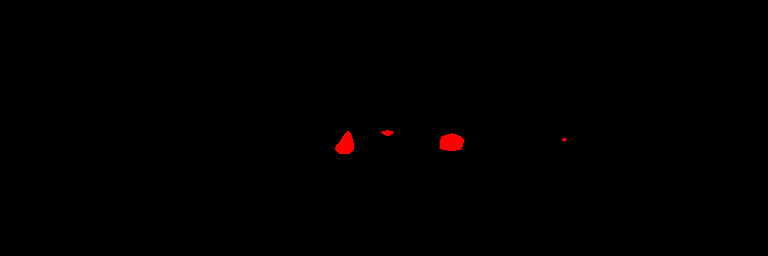} &
\adjustimage{height=1.12cm,valign=m}{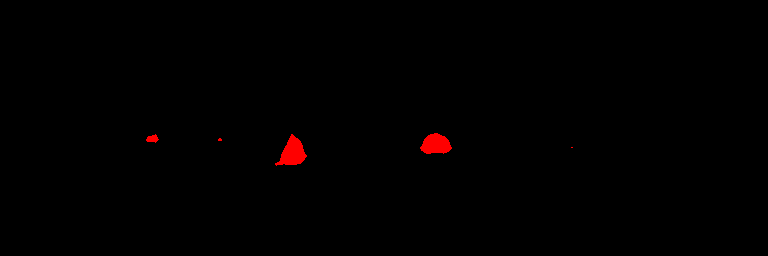} &
\adjustimage{height=1.12cm,valign=m}{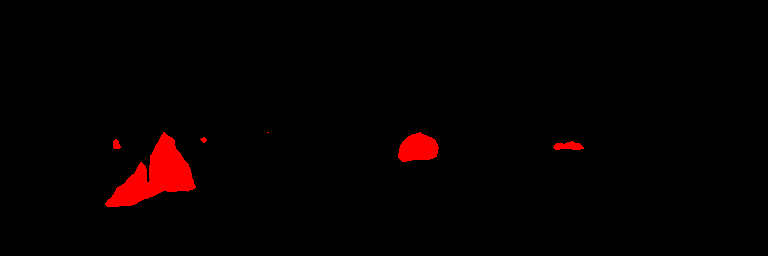} &
\adjustimage{height=1.12cm,valign=m}{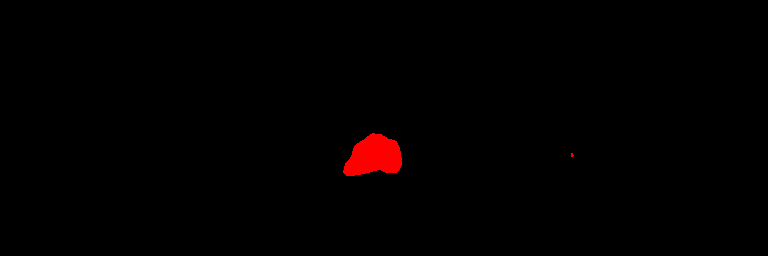} \\

\end{tabular}
\captionof{figure}{
Qualitative localization results trained with different pseudo masks. 
Rows from top to bottom: camera frames, ground-truth masks, predictions without pseudo mask selection, and predictions with pseudo mask selection.
}
\label{fig:loc_qual}
\end{table*}

\subsubsection{Localization Model Performance}
\label{sec:localization}
We evaluate our localization model with other weakly-supervised baselines. As shown in Table~\ref{tab:localization-performance}, the MIL-based approach~\cite{chen2023weakly} performs the worst across all spatial-temporal metrics. 
This indicates that MIL loss alone is insufficient for localizing atomic activities under weak supervision.
In contrast, incorporating the contrastive loss from VSCR~\cite{duan2024mining} yields only marginal improvements over our base model (22.8\% vs. 22.7\% mIoU), which is trained on initial pseudo masks without selection. 
This is because VSCR's contrastive objective heavily relies on appearance cues, which offer limited discrimination for atomic activities where actors share highly similar, overlapped visual features. 
Moreover, without a mechanism to filter out spatial noise, pulling these unrefined regions closer in the feature space can mistakenly entangle the true target with background distractors, leaving the model vulnerable to spurious attention responses.
Overall, our localization model, explicitly trained using pseudo masks refined by the selection stage, achieves the best performance. This validates the effectiveness of our attention difference selection strategy: by leveraging counterfactual intervention to eliminate spurious attention responses, it yields highly precise pseudo-supervision for downstream localization training.

Qualitative examples in Figure~\ref{fig:loc_qual} illustrate localization results of querying different atomic activities in cluttered traffic scenes with multiple co-occurring activities. 
Without the selection stage, initial predictions roughly capture the regions of interest but often include irrelevant areas, leading to significant false positives.
With the refined pseudo masks fine-tuning, localization becomes more precise, with improved boundary alignment and more effective suppression of irrelevant regions.

These improvements are particularly evident in cases where multiple agents perform similar activities. As illustrated in Figure~\ref{fig:loc_qual}, when querying \textbf{Z2-Z1:K+}, the selection stage enables the model to correctly distinguish between two bicycles and a red van driving the same pattern, whereas the initial predictions erroneously include the van in the mask. 
Such results highlight the advantage of our selection framework in producing sharper and more discriminative localization.

\subsubsection{Robustness of Absent-Action Queries}
\label{sec:fp}

We assess the false positive rate when querying atomic activity categories that are absent in videos.
%
For each video in the testing set, we randomly sample a query of an absent actor category (e.g., pedestrian).
%
A sample is treated as a false positive if the number of activated pixels is larger than a threshold. 
To account for varying object sizes, we apply actor-type-specific thresholds on the number of activated pixels. Specifically, we group the actions into three main categories based on the actor type: vehicle (C and C+), two-wheeler (K and K+), and pedestrian (P and P+). 
The threshold for each actor type is determined by the average number of pixels in a video, calculated from the ground truth annotations. 
Our results reported in Table~\ref{tab:fp-rate} show that the model trained with refined pseudo masks exhibits a substantial reduction in false positives across all actor types, even without any explicit optimization for handling non-present actions. 
Importantly, this improvement is not simply due to globally suppressing activations during the process of identifying important regions at the selection stage; when querying present activities, the selection stage increases recall and IoU (Table~\ref{tab:localization-performance}), demonstrating that it selectively removes spurious activations while preserving true activity coverage.

\begin{table}[t!]
\centering
\caption{Ablation study on attention-object matching strategies and comparison with the overlap-based approach by Chen \textit{et al.}~\cite{chen2020learning}.}
\resizebox{\columnwidth}{!}{%
\begin{tabular}{lcccccc|c}
\toprule
\textbf{Method} & \textbf{C} & \textbf{C+} & \textbf{K} & \textbf{K+} & \textbf{P} & \textbf{P+} & \textbf{mIoU} \\
\midrule
Overlap-based~\cite{chen2020learning} & 1.5 & 1.4 & 0.7 & 0.5 & 0.2 & 0.4 & 0.9 \\
Distance & 31.7 & 27.3 & \textbf{29.9} & 22.3 & \textbf{23.6} & \textbf{12.3} & 25.8 \\
Gravity & \textbf{41.7} & \textbf{27.4} & 28.4 & 24.3 & 17.7 & 11.3 & 27.1 \\
Mixed & \textbf{41.7} & \textbf{27.4} & \textbf{29.9} & 22.3 & \textbf{23.6} & \textbf{12.3} & \textbf{28.9} \\
\bottomrule
\end{tabular}%
}
\label{tab:ablation-matching}
\end{table}

\subsubsection{Ablation Study}\hfill\break
\label{sec:loc_ablation}
\noindent \textbf{Attention-Object Matching.} 
We ablate the attention-object matching strategy by comparing our mixture matching introduced in Sec.~\ref{sec:matching} with the overlap-based pseudo mask generating algorithm~\cite{chen2020learning}.
%
As shown in Table~\ref{tab:ablation-matching}, this threshold-and-overlap paradigm fails drastically in our experiment. 
The primary cause of this failure is its heavy reliance on spatial overlap with an initial binary mask $M_{\text{init}}$. 
In complex driving scenes, applying hard thresholding to continuous attention heatmaps inevitably yields a highly noisy, fragmented, or overly dilated $M_{\text{init}}$. 
When the initial mask is severely degraded, the overlap-based matching criterion completely breaks down---yielding near-zero overlap for valid small actors or erroneously merging adjacent distractors. 
In contrast, our proposed matching metrics bypass this fragile pixel-level overlap. Interestingly, we observe that the Gravity and Distance metrics perform comparably on two-wheelers (K and K+). 
We attribute this to the intermediate and highly variable relative scale of two-wheelers in real-world traffic: they act as smaller objects when compared to four-wheelers, but as larger objects when compared to pedestrians. This dual nature of their relative scale causes both distance-based and area-weighted (Gravity) metrics to be similarly effective.
Overall, by leveraging the robust geometric properties (centers and areas) rather than fragile overlap conditions, our proposed \textbf{mixture matching} approach yields the best average performance across all categories in terms of mIoU and justifies our design choice.

\begin{table}[t!]
\centering
\caption{Ablation study on query representations of the atomic activity for localization.
Selection denotes whether the mask selection mechanism is applied.}\begin{tabular}{lccccc}
\toprule
\textbf{Input}&  \textbf{Selection}&\textbf{mIoU}& \textbf{oIoU}& \textbf{tIoU}& \textbf{mAP@tIoU}\\
\midrule
One-hot& \multirow{3}{*}{\xmark} & 11.7& 15.0& 7.5& 0.8\\
Slot features& & 12.7& 13.8& 7.9& 0.9\\
Attention& & 22.7& 25.7& 20.3& 5.7\\
\midrule
One-hot&  \multirow{3}{*}{\cmark}& 12.4& 13.6& 7.2& 0.7\\
Slot features&  & 13.7& 14.0& 9.0& 1.1\\
Attention&  & \textbf{24.7}& \textbf{28.1}& \textbf{23.5}& \textbf{6.8}\\
\bottomrule
\end{tabular}
\label{tab:ablation-query}
\end{table}

\noindent \textbf{Query Representation.}
We study the query representation by comparing conventional one-hot class embeddings and the Action-Slot's representations---class-specified slot features and the corresponding attention.
%
Table~\ref{tab:ablation-query} shows that one-hot embeddings and slot features fail to provide sufficient hints for localization, yielding only 12.4\% and 13.7\% mIoU, respectively. 
In contrast, using the attention map significantly improves the performance to 24.7\%.
This indicates that attention maps, inherently derived from feature-level dot-products, retain much richer spatial cues about the target activity. 
While slot features encode action semantics, they can barely recover the spatial-temporal cues.
Finally, we observe that applying our \textit{selection} mechanism (detailed in Sec.~\ref{sec:selection}) consistently improves the performance across all representations by filtering out irrelevant objects via counterfactual intervention.

\begin{table*}[t!]
\begin{center}

\begin{tabular}{c@{\;}c@{\;}c@{\;}c@{\;}c@{\;}c}
\hspace{-4mm}
& \adjustimage{trim=35mm 5mm 45mm 10mm, clip, height=2cm,valign=m}{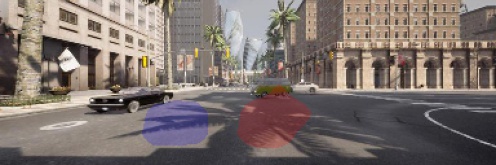}
& \adjustimage{trim=35mm 5mm 45mm 10mm, clip, height=2cm,valign=m}{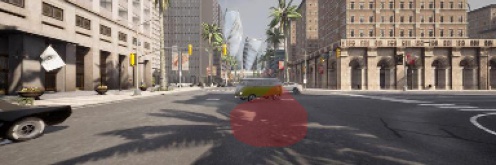}
& \adjustimage{trim=35mm 5mm 45mm 10mm, clip, height=2cm,valign=m}{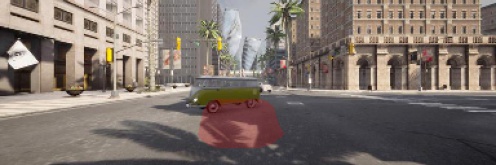}
& \adjustimage{trim=35mm 5mm 45mm 10mm, clip, height=2cm,valign=m}{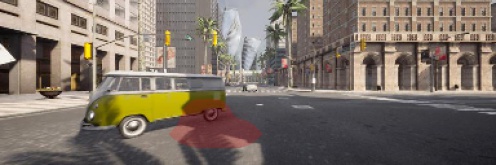}
& \adjustimage{height=1.9cm,valign=m}{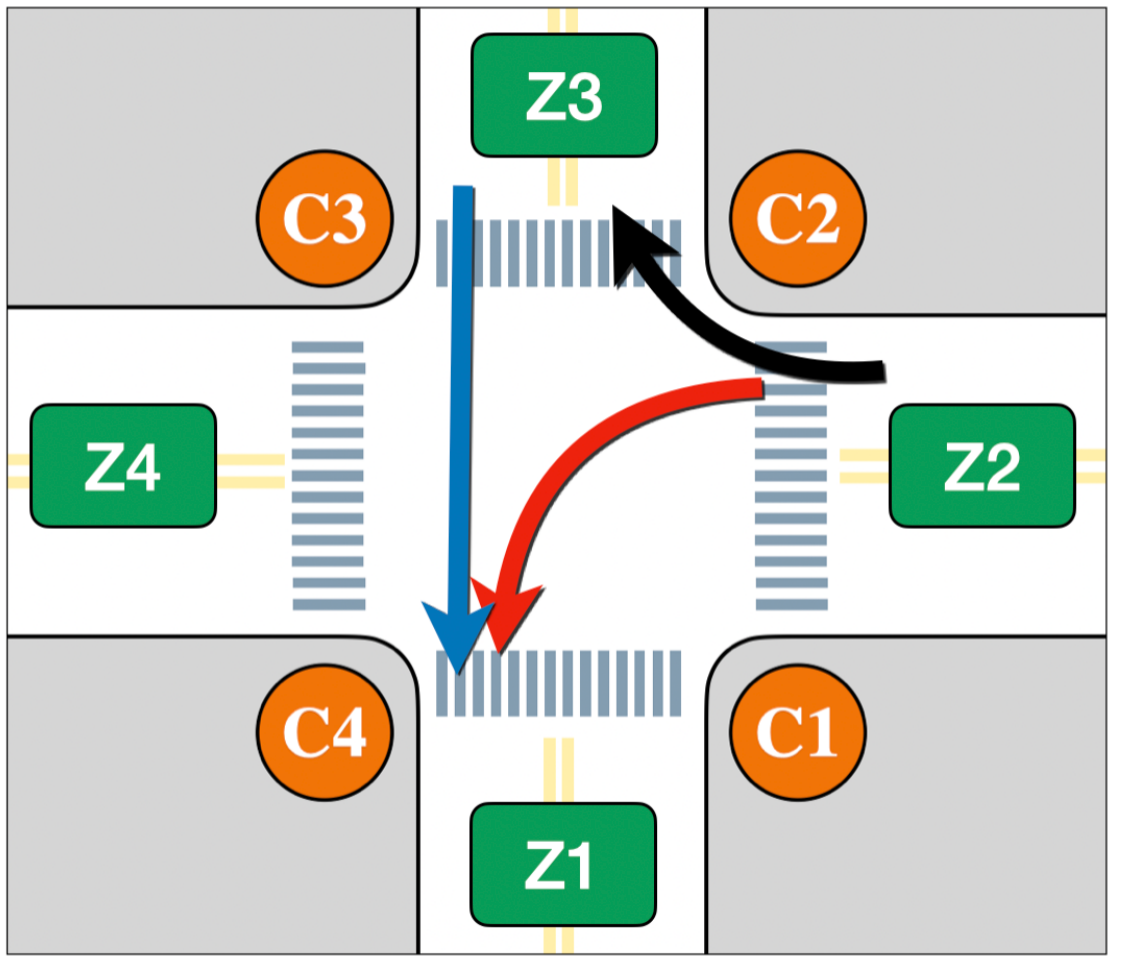}

\end{tabular}
\end{center}
\captionof{figure}{
Action-Slot attention visualization in a failure case involving occlusion. 
Distinct mask colors represent attention maps for different atomic activity classes. 
The yellow bus partially occludes the white car emerging from the right roadway (Z2). Action-Slot correctly predicts \textcolor{blue}{Z3-Z1:C} and \textcolor{red}{Z2-Z1:C}, but fails to detect \emph{Z2-Z3:C}, which is performed by the occluded white car (indicated by the black arrow).}
\label{fig:occlusion}
\end{table*}

\noindent \textbf{Input Features.}
We examine whether Action-Slot provides stronger input representations for localization by comparing them with features from video models. 
As shown in Table~\ref{tab:ablation-features}, using Action-Slot features yields substantial overall improvements compared to their raw backbone counterparts. 
This demonstrates that Action-Slot learns more structured, action-centric representations that are significantly more informative for atomic activity localization than standard video backbones.

\begin{table}[t!]
\centering
\caption{Ablation study on input features for localization, comparing raw features from standard video backbones against those enhanced by the Action-Slot.}
\begin{tabular}{lcccc}
\toprule
\textbf{Input Features}& \textbf{mIoU}& \textbf{oIoU}& \textbf{tIoU}& \textbf{mAP@tIoU}\\
\midrule 
I3D~\cite{carreira2017quo}& 13.2& 15.4& 11.5& 2.5\\
I3D w/ Action-Slot& 13.8& 19.9& 12.5& 3.5\\
\midrule
X3D~\cite{feichtenhofer2020x3d}& 14.3& 15.2& 12.9& 2.2\\
X3D w/ Action-Slot& \textbf{24.7}& \textbf{28.1}& \textbf{23.5}& \textbf{6.8}\\
\bottomrule
\end{tabular}
\label{tab:ablation-features}
\end{table}

\section{Conclusions}

\label{sec:conclusion}
In this work, we present a structured action-centric representation learning framework that reformulates slot attention for multi-agent atomic activity understanding.
%
%
We show that the learned action-centric representations serve as a unified foundation for two complementary aspects of atomic activity understanding: multi-label recognition and spatio-temporal localization.
Extensive experiments demonstrate that Action-slot achieves state-of-the-art performance on OATS and the proposed TACO dataset, while significantly improving cross-domain generalization through balanced, topology-aware pretraining.
Beyond recognition, we demonstrate that the structured slot representations encode meaningful spatial grounding signals. Leveraging these signals, we develop a weakly supervised localization framework that refines slot attention into high-quality pseudo masks. The resulting model achieves superior localization accuracy and substantially reduces false positives under absent queries, without requiring dense annotations.
%
%

%
Overall, our findings suggest that action-centric structured representations provide a principled and transferable foundation for multi-agent video understanding. By bridging recognition and localization within a unified representation paradigm, this work validates the potential of structured decomposition as a scalable alternative to annotation-intensive approaches for complex dynamic scenes. More broadly, this unified perspective opens a new direction for structured video representation learning, where semantically aligned decomposition enables scalable recognition and localization under weak supervision.

\noindent \textbf{Limitation.}
Although Action-Slot demonstrates favorable quantitative and qualitative performance, we observe that it is still challenging to handle cases when two activities occlude with each other, as shown in Figure~\ref{fig:occlusion}.
In these cases, the corresponding action slots may be confused about where they should attend.
This observation is closely relevant to the tracking task under occlusion, where frequent ID switches are often observed.
%
Moreover, the evaluation of atomic activity localization still relies heavily on manually labeled data.
To advance the field, we invite the community to explore automatic or semi-automatic labeling strategies that can enable more comprehensive and efficient evaluation.


\appendices

\bibliographystyle{IEEEtran}
\bibliography{main}


\end{document}